\documentclass[10pt]{article} % For LaTeX2e
\usepackage[accepted]{tmlr}
\usepackage{hyperref}
\usepackage{url}
\usepackage{xspace}
\usepackage{amsmath}
\usepackage{amsthm}
\usepackage{amsfonts}
\usepackage{mathtools}
\usepackage{bm}
\usepackage{algorithm}

\let\AND\relax
\usepackage{algorithmic}
\usepackage{multirow}
\usepackage{dsfont}
\usepackage{pict2e}
\usepackage{array}
\usepackage{mathtools}

\usepackage{microtype}
\usepackage{graphicx}
\usepackage{booktabs} % for professional tables

\usepackage{color, colortbl}
\definecolor{Gray}{gray}{0.9}

\usepackage{subfig}
\usepackage{caption}
\usepackage{makecell}
\usepackage{balance}

\usepackage{array}
\usepackage{enumitem}
\usepackage{pifont}% http://ctan.org/pkg/pifont
\usepackage{tabu}
\usepackage{tablefootnote}

\newcommand{\ks}[1]{\textcolor{red}{[KS: #1]}}

\DeclareMathOperator*{\argmin}{arg\,min}

\newcommand{\ie}{\textit{i.e.}\xspace}
\newcommand{\eg}{\textit{e.g.}\xspace}

\newcommand{\hbf}{\mathbf{h}\xspace}

\newcommand{\Xbf}{\mathbf{X}\xspace}

\newcommand{\bbf}{\mathbf{b}\xspace}

\newcommand{\gbf}{\mathbf{g}\xspace}

\newcommand{\zerobf}{\mathbf{0}\xspace}

\newcommand{\Gbf}{\mathbf{G}\xspace}

\newcommand{\Kbf}{\mathbf{K}\xspace}

\newcommand{\Wbf}{\mathbf{W}\xspace}

\newcommand{\mubf}{\bm{\mu}\xspace}

\newcommand{\Thetabf}{\bm{\Theta}\xspace}
\newcommand{\Phibf}{\bm{\Phi}\xspace}
\newcommand{\one}{\mathds{1}\xspace}

\newcommand{\CV}{\mathcal{V}\xspace}
\newcommand{\CE}{\mathcal{E}\xspace}

\newcommand{\CL}{\mathcal{L}\xspace}

\newcommand{\CY}{\mathcal{Y}\xspace} 
\newcommand{\CN}{\mathcal{N}\xspace}

\newcolumntype{H}{>{\setbox0=\hbox\bgroup}c<{\egroup}@{}}

\newtheorem{problem}{Problem}

\title{Personalized Layer Selection for Graph Neural Networks
}
\author{\name Kartik Sharma~\thanks{Work done during internship at Visa Research} \email ksartik@gatech.edu \\
      \addr Georgia Institute of Technology 
      \AND \\ \\
      \name Vineeth Rakesh \email vinmohan@visa.com \\
      \addr Visa Research 
      \AND \\ \\
      \name Yingtong Dou \email yidou@visa.com \\
      \addr Visa Research
      \AND \\ \\
      \name Srijan Kumar \email srijan@gatech.edu \\
      \addr Georgia Institute of Technology
      \AND \\ \\
      \name Mahashweta Das \email mahdas@visa.com \\
      \addr Visa Research}

\def\month{04}  % Insert correct month for camera-ready version
\def\year{2025} % Insert correct year for camera-ready version
\def\openreview{\url{https://openreview.net/forum?id=JyjTJAG9yZ}} % Insert correct link to OpenReview for camera-ready version

\begin{document}

\maketitle

\begin{abstract}
    % Graph-structured data is ubiquitous in various domains such as finance, biology, social media, and e-commerce. In the literature, /scope
    Graph Neural Networks (GNNs) combine node attributes over a fixed granularity of the local graph structure around a node to predict its label. However, different nodes may relate to a node-level property with a different granularity of its local neighborhood, and using the same level of smoothing for all nodes can be detrimental to their classification. In this work, we challenge the common fact that a single GNN layer can classify all nodes of a graph by training GNNs with a distinct \textit{personalized} layer for each node. Inspired by metric learning, we propose a novel algorithm, \textbf{MetSelect}, to select the optimal representation layer to classify each node. 
    % train off-the-shelf GNN classifiers with personalized scope using a metric learning training strategy. 
    In particular, we identify a prototype 
    % mean \yd{prototype} 
    representation of each class in a transformed GNN layer and then classify each node using the layer where the distance is smallest to a class prototype after normalizing with that layer's variance. Results on $10$ 
    datasets 
    % 6 heterophilic datasets \yd{mention homophilic data for comprehensiveness} 
    and $3$ different GNNs show that we significantly improve the node classification accuracy of GNNs in a plug-and-play manner. We also find that using variable layers for prediction enables GNNs to be deeper and more robust to poisoning attacks.
    % to converge faster (in as little as $5$ epochs), be deeper (as high as $128$ layers), heterophily-resistant (up to $25\%$ average gain), robust ($30\%$ better against poisoning attacks)
    % , and to identify the scope around the node that can explain its prediction. 
    We hope this work can inspire future works to learn more adaptive and personalized graph representations. 

    % \ks{Existing local smoothing approach is not dataset and model dependent}
\end{abstract}

\section{Introduction}\label{sec:intro}
% {\color{red}[VR: While there is no major mistakes in the introduction section, I feel that this section could be improved. The motivating examples such as pyramid scheme, etc are very abstract. People who are not from the same domain might not be able to get the motivation as to why different views are needed. While Figure 1 does convey the strengths of our work, I feel it needs a bit more work to motivate the readers. ]}\ks{Check now.}

\begin{figure}[t]
    \centering
    \includegraphics[width=0.9\textwidth]{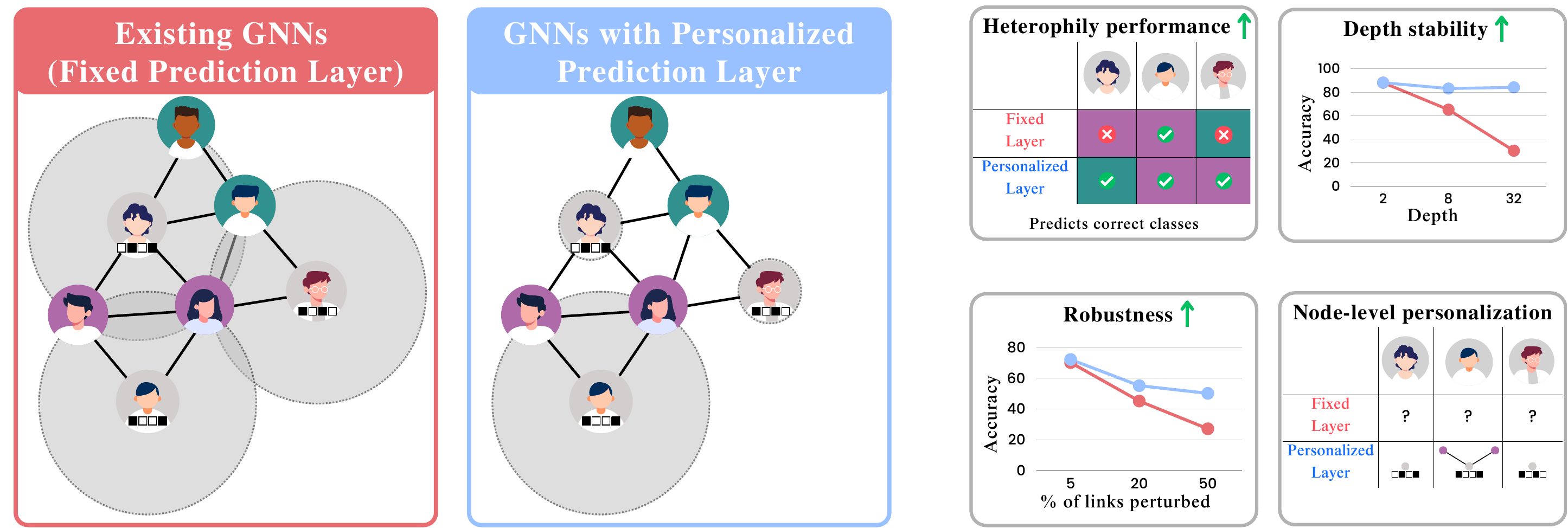}
    \caption{GNNs with personalized prediction layer can empower existing GNNs for more \textit{accurate}, \textit{personalized}, \textit{robust}, and \textit{deeper} node classification (\eg, classifying unlabeled users (gray) as conservative (purple) or liberal (green)). For example, while some nodes benefit from their neighbors, others may be better off by predicting using only their attributes. 
    % \ks{update}
    % GNNs with personalized scopes improve heterophily performance, depth, robustness, with explainable scopes. \ks{remove explainable}} 
    }
    \label{fig:problem}
\end{figure}

Graph Neural Networks (GNNs)~\citep{hamilton2017inductive,kipf2016semi} show superior performance on various node-level and graph-level tasks for real-world applications, including fraud detection~\citep{tang2022rethinking} %dou2020enhancing,
% \yd{replace my paper to this one: Rethinking Graph Neural Networks for Anomaly Detection}
% , molecular property prediction~\citep{wang2022molecular}, 
and online content recommendation~\citep{ying2018graph}
% ,sharma2022survey}
. 
% They exploit the node's attributes and local graph structure to predict its label, \eg, whether a user is conservative or liberal in a social network. 
A GNN of depth $L$ iteratively learns $L$ different representations by combining a node's attributes with its local neighborhood~\cite{gilmer2017neural} and then use the final layer representation to classify each node. However, nodes in a graph differ from each other and their underlying property may correlate with a distinct granularity of each node's neighborhood.
% local neighborhood. 
% and each node may have a an underlying property may correlate with a distinct scope of its local neighborhood. 
% combines a node's attributes with its neighbors for $L$ steps, incorporating information from the $L$-hop neighbors of each node~\cite{gilmer2017neural}. However, using the same \emph{scope}, \ie, the number of hops to classify each node can be suboptimal. 
% A node's property may correlate with a distinct scope of its local neighborhood.
% Each node has a distinct way of combining with its neighborhood to generate its label. 
% For example, for fraud detection (commonly regarded as a binary classification task), while one can detect an identity theft using just the node's attributes, detecting a pyramid scheme fraud may require higher-order transactional activity. [{\color{red} VR: Need a better example and Figure 1 should be designed to guide this example}] Here, using the same scope to encode both fraud nodes can be suboptimal since a lower scope/depth may remove useful higher-order information, while a higher scope/depth can lead to \textit{oversmoothing} of the individual attributes due to possibly dissimilar/\textit{heterophilic} connections~\citep{bodnar2022neural,nt2019revisiting}. 
For instance, consider the task of predicting the political ideologies (\ie, conservative or liberal) of users in a social network. Users tend to form echo chambers by interacting with people of similar ideologies~\citep{cinelli2021echo}, whereas some doubtful users may interact with diverse ideologies while leaning towards one~\citep{boutyline2017social}. 
While the neighbor information may be useful to accurately classify the users within an echo chamber, it can confuse a classifier for the users that interact with both chambers where using just the users' traits may be better. 
% While a GNN of depth $1$ can exploit the homophilous nature of the users within an echo chamber, it may smoothen the diverse ideologies of the neighbors of a node that interacts with both chambers, leading to inaccurate classification. 
Figure~\ref{fig:problem} highlights the importance of using a personalized layer instead of a fixed layer of a GNN with this example. 

We hypothesize that selecting a personalized GNN layer has wide-ranging implications. 
Existing GNNs are known to suffer from the problem of oversmoothing, where node representations become indistinguishable due to convolution~\citep{nt2019revisiting}. This leads to poor performance on datasets where labels are typically associated with lower-level features. 
Existing works have focused on handling this issue by either enhancing the representations with higher frequency channels~\citep{luan2022revisiting,chien2020adaptive,nt2019revisiting,eliasof2023improving,zhao2019pairnorm}, continuous neural diffusion processes~\citep{bodnar2022neural,di2022graph}, or by rewiring the underlying graph~\citep{rong2019dropedge,gutteridge2023drew}. 
However, we note that one can \textbf{enhance the accuracy} of existing GNNs by simply using a node-personalized representation for prediction instead of the final representation. 
% Existing works have proposed novel architectures that exploit information from higher frequency channels or that use continuous neural diffusion processes~\citep{bodnar2022neural,di2022graph}, and random structural transformations~\citep{rong2019dropedge} to handle oversmoothing in GNNs. However, none of these approaches attempt to handle 
% perform poorly on heterophilic datasets, even when simple Multi-Layer Perceptrons (MLPs) show high performance~\citep{lim2021large}. This means that the node attributes themselves are enough to classify the nodes well but these features get smoothened due to their neighbors~\citep{bodnar2022neural}.  
% A flexible neighborhood scope can ignore the irrelevant neighbors in either higher or lower-level neighborhoods. 
We can also circumvent the problem of loss in performance in \textbf{deeper} GNNs~\citep{li2018deeper} by identifying the optimal layer for each node and thus, ignoring the node representations at higher depths if they don't help to classify that node. 
Using personalized GNN layer also boosts their \textbf{robustness} to untargeted training-time perturbations~\citep{zugner2018adversarial,sun2020non,xu2019topology}. These perturbations are carefully crafted to reduce the overall accuracy of the GNNs over a given dataset but their effect will be limited since we can adaptively select the layer that is useful for classification and ignore any redundant information that may be perturbed. 
% ignore any redundant information outside that scope for each node. 
% {\textcolor{red}{Furthermore, metric learning enables enhanced robustness since normalized distance to class mean representations is likely to be less affected than a softmax-based prediction~\citep{mao2019metric}.  }}
Finally, these improvements can be provided to any GNN in a \textbf{plug-and-play} manner while preserving their analytic simplicity by \textbf{personalizing} the selection of these representations. 

% Problem~\ref{prob:main} assumes the GNN model to be given and thus, we do not propose a new encoder. This means we can achieve the enhancements of a variable neighborhood scope in a plug-and-play manner for any underlying GNN architecture. This removes the burden of design choice and innovation. 

% \textbf{Enables deeper architectures.} 

% \textbf{Ignores heterophilic neighbors.} 

% \textbf{More robustness.} 

% \textbf{Faster convergence.} 

% \textbf{Plug-and-play.} 

% While no existing works have focused on the problem of empowering GNNs with different neighborhood scopes, they have focused on other related issues. Graph Structure Learning~\citep{zeng2021decoupling,jin2020graph} aims to extract the optimal graph structure for the underlying task but this makes the predictions obscure as the graph structure is no longer preserved for the prediction as in our problem. Oversmoothing issues have also been dealt with by proposing novel architectures that exploit information from higher frequency channels~\citep{luan2022revisiting,chien2020adaptive,nt2019revisiting} or consider continuous neural diffusion processes~\citep{bodnar2022neural,di2022graph}. However, we focus on providing a plug-and-play improvement over existing GNNs for oversmoothing by just using a variable scope for each node's neighborhood. 

In this work, we first formalize the problem of finding the optimal GNN layer for each node to solve the transductive node classification with a given GNN. Then, we propose a novel method, called \textbf{MetSelect} that leverages \textbf{Met}ric learning to \textbf{select} the optimal layer in an efficient manner. Here, we find a prototype representation of each class in transformations of each GNN layer and then compare the metric distance of a node to these prototypes across layers to find the optimal layer. To train the models, we minimize the cross entropy loss of the decoded probabilities from the optimal layers for each training node while minimizing their distance to the true class prototype in the transformed space. Through experiments on $10$ datasets and $3$ GNNs, we show that MetSelect boosts the node classification performance by up to $20\%$, especially on heterophilic datasets. We also find that MetSelect enables deeper GNNs by preserving the accuracies for as high as $10$ layers. GNNs trained using MetSelect are also found to be robust against untargeted poisoning attacks~\citep{zugner2018adversarial}, giving up to $100\%$ enhancement for different graph neural networks. 
\section{Related Work}\label{sec:related_work}

\paragraph{Layer aggregation.} Layer aggregation methods such as JK-net, try to learn personalized representations by aggregating different representations through pooling operations and/or skip connections for node classification~\citep{xu2018representation,dwivedi2023benchmarking,rusch2022gradient}. However, they do not adaptively select a single layer for classifying a given node but rather merge the representations for all nodes in the same way. This reduces the interpretability and increases the complexity of the underlying network by predicting through non-linear pooled transformation of the GNN representations. On the other hand, the simpler layer selection of MetSelect preserves the interpretability and the analytic simplicity of the underlying GNN as the predictions are still made through the GNN embeddings. Furthermore, layer aggregation methods require all representations to lie in the same space and to have the same dimension while selecting optimal layers through MetSelect completely relaxes such assumptions. 

\paragraph{Alternative GNN architectures.} 
% An alternative view of graph neural networks is provided by graph signal processing~\citep{ortega2018graph,dong2020graph} where graph filters are learned that transform the graph Laplacian spectrum into embeddings. 
% It has been shown that common GNNs suffer from oversmoothing and poor performance on heterophilic datasets due to them learning a low-pass filter~\citep{nt2019revisiting,di2022graph,yan2022two}, \ie, they only encode the low-frequency information. These observations have led the researchers to propose 
Based on the observations that a GCN can only approximate a low-pass filter~\citep{nt2019revisiting,di2022graph,yan2022two}, new models have been proposed to address the heterophilic~\citep{pei2020geom,luan2022revisiting,eliasof2023improving,chien2020adaptive,rossi2024edge} and closely-related oversmoothing~\citep{bodnar2022neural,yan2022two} challenge. However, these models achieve these enhancements by modifying the underlying message-passing architecture, while we present a plug-and-play approach to show that these issues can possibly be addressed by adaptive layer selection. NDLS~\citep{zhang2021node} also finds node-specific smoothing level, their method is based on a simple feature smoothing model and is agnostic of the underlying GNN and the label distribution. Furthermore, while they hypothesize that the optimal layer is the one at which the influence of other nodes is similar to the stationary influence, we hypothesize that given a GNN, it is the one that minimizes the distance from the class prototype in the corresponding embedding space.

% While these models can incorporate information from multiple channels, they do this at the expense of  information aggregation of each node by moving away from the inductive bias of the message-passing architecture. In contrast, we focus on the problem of selecting the best layer for each node to improve the performance of the base GNNs without making any architectural changes. 

\paragraph{Alternative training of GNNs.} Alternative training strategies have been proposed to support larger graphs~\citep{gandhi2021p3}, faster training~\citep{cai2021graphnorm}, longer distances~\citep{li2021training,gutteridge2023drew,rong2019dropedge}, robustness~\citep{gosch2023adversarial}, and a joint structure and embedding learning~\citep{jin2020graph,zeng2021decoupling}. However, none of these approaches focus on our problem of selecting the optimal personalized layer and instead focus on increasing either robustness or faster convergence. \citet{zeng2021decoupling} decouples the depth and the scope of the GNNs to consider a neighbor multiple times in the message passing by using a GNN of depth greater than the neighborhood scope of that node. On the other hand, our objective here is to find an optimal representation depth $L’$ strictly less than the full scope $L$ that best predicts the node's class.

\paragraph{Neighbor importance.} Certain methods aim to identify useful neighbors for each node separately to better classify them.
% Explanations can be either through a \textit{post-hoc} explainer module~\citep{ying2019gnnexplainer,schnake2021higher,huang2022graphlime} or by training \textit{self-interpretable} GNNs that can explain their own predictions~\citep{zhang2022protgnn,feng2022kergnns,dai2021towards}. 
% These approaches tend to focus on graph-level tasks due to their applications in the molecular domain of finding an important group for classification~\citep{kakkad2023survey}. 
% On the other hand, our approach shows that explainable predictions can be provided without explicitly changing its architecture by simply changing how the classes are decoded.
% is a novel attempt to enhance the interpretability of node-level representations of existing GNNs. 
Self-interpretable~\citep{dai2021towards,zhang2022protgnn,feng2022kergnns} and attention-based~\citep{velivckovic2017graph,shi2020masked,brody2021attentive} graph neural networks weigh nodes based on the similarity between different nodes in the whole graph or the local graph structure. While this implicitly weighs each neighbor differently, the representations of all nodes are still formed from a fixed number of message-passing steps. Thus, these are still susceptible to oversmoothing and are not scalable as opposed to their graph neural network counterparts. Instead, we provide plug-and-play improvements to out-of-the-shelf GNNs by simply presenting a way to select the optimal smoothing level for each node. 

\paragraph{Metric Learning.} Our strategy to choose the optimal layer is inspired by prior works in the image domain to learn embedding spaces by forming multiple clusters per class~\citep{rippel2015metric,deng2020sub}. While it is prevalent to use distance-based classifiers in images~\citep{deng2019arcface,salakhutdinov2007learning} for learning robust embeddings, they have not been explored in the context of graph neural networks before. In this work, we use the embedding distance from class clusters to do personalized node-level smoothing.
% with provides us with a notion of multiple prototypes per class as the class means $\mubf^{(l)}_c$. These denote the prototype representations at different scope levels of each class which further enhance the few-shot adaptability~\citep{allen2019infinite,snell2017prototypical} and interpretability~\citep{zhang2022protgnn}. Prototypical networks are effective in few-shot learning~\citep{allen2019infinite} and by mapping the prototypes back to a structure, we can understand more about what the model has learned about the class~\citep{zhang2022protgnn}. However, since this does not pertain to Problem~\ref{prob:main}, we leave it for future works to explore these aspects of our method in more detail. 

\section{Background}

In this work, we study the problem of \textit{transductive node classification}, where we are given a partially labeled graph and we want to learn the labels of the remaining nodes. More formally, 
% it can be defined as

\begin{problem}{\textbf{(Transductive Node Classification)}}
    Given a graph $\Gbf = (\Xbf, (\CV, \CE))$, with node attributes $\Xbf$, nodes $\CV$, edges $\CE$, and a set of labeled nodes $\{(v, y): v \in \CV_{tr} \subset \CV, y \in \CY\}$, the objective is to learn a function $f_{\Wbf}(\cdot)$ to predict the label $y \in \CY$ of each node $v \in \CV$. 
\end{problem}

% \begin{problem}{\textbf{(Self-interpretable Node Classification)}}
%     Given a graph $\Gbf = (\Xbf, (\CV, \CE))$ and a set of labeled nodes, the objective is to learn a function $f_{\Thetabf}(\cdot)$ that can (1) predict the label $y \in \CY$ of each node $v \in \CV$, and (2) provide a local explanation $\Gbf_{v,y}$ from the ego-network of $v$. 
% \end{problem}

We then consider an encoder-decoder architecture of $f_{\Wbf}(\cdot) := \gbf_{\Phi} \circ \hbf_{\Theta}(\cdot) \in [0, 1]^{|\CY|}$, that predicts the probability of belonging to each class using a (trainable) decoder function $\gbf_{\Phi}$. The encoder $\hbf_{\Theta}$ is a message-passing Graph Neural Network (GNN)~\cite{kipf2016semi,gilmer2017neural} of depth $L$ that encodes each node $v_i$ as $\hbf_{\Theta}(v_i) := \hbf^{(L)}_i \in \mathbb{R}^d$ defined recursively as: 
% $\hbf^{(l+1)}_i = \text{Update}(\hbf^{(l)}_i, \text{Agg} (\{\text{msg}(\hbf^{(l)}_j, \hbf^{(l)}_i): (v_j, v_i) \in \CE\}))$,
\begin{equation}\label{eq:gnn}
    % \resizebox{.9\hsize}{!}{%
    % $
    \hbf^{(l+1)}_i = \text{Update}(\hbf^{(l)}_i, \text{Agg} (\{\text{msg}(\hbf^{(l)}_j, \hbf^{(l)}_i): (v_j, v_i) \in \CE\})),
    % $,%
    % }
\end{equation}
where $\text{Update}(\cdot), \text{Agg}(\cdot), \text{msg}(\cdot)$ are learnable functions, $\hbf^{(0)} = \Xbf$, and the $l$-th \textit{\textbf{layer}} $\hbf^{(l)}$ is a node representation formed using the neighbors at a distance of at most $l$ \textit{\textbf{hops}} from that node, \ie, $\CN^{(l)}(v_i)$. 

\section{Problem}\label{sec:problem}
% \ks{Give a better example} 
GNNs aggregate the $L$-hop information of each node to learn their label-separating representations. However, a major limitation stems from the fact that the same neighborhood scope $L$ is used to classify all the nodes. As shown in Figure~\ref{fig:problem}, different nodes may require a different granularity of the local information to be classified properly, and providing more information could unnecessarily smooth out the classifying signal.
To predict the user ideologies in a social network, nodes within an echo chamber may benefit from their $1$-hop neighbors. However, a classifier may be better off ignoring the neighbors of the nodes that interact with both chambers and just using the attributes ($0$-hop).
% This is highlighted in a binary classification fraud detection problem as a pyramid-scheme fraud (1-hops) is different from an identity-theft fraud (0-hops). 
% For example, suppose we want to detect fraudulent accounts from a transactional network. While an identity theft can be identified using the attributes itself ($0$-hop), a pyramid scheme fraud requires more than $1$ hops to be detected. Thus, using the same number of layers for both nodes may give suboptimal performance. 
To capture such intraclass differences, we propose a novel problem that aims to classify a node using the optimal layer for each node. In other words, 
\begin{problem}{\textbf{(Node-personalized layer selection in GNN)}}\label{prob:main}
    Given a graph $\Gbf = (\Xbf, (\CV, \CE))$ and a GNN encoder $\hbf_{\Theta}(\cdot)$ of depth $L$, the objective is to identify a personalized layer $l^*(v) \in [0, L]$ for each node $v \in \CV$ such that $\hbf^{(l^*(v))}(v)$ best predicts the true label $y(v)$. 
\end{problem}

% \textcolor{red}{
% Building on this motivation, certain nodes may be more accurately classified using representations learned by one encoder over another. Thus, we also study the problem of selecting the most suitable graph neural network for each node from a fixed set of encoders to optimize prediction performance. For example, suppose we want to choose between an MLP and a GCN to classify each node. More formally,
% \begin{problem}{\textbf{(Selecting node-optimal GNNs)}}\label{prob:main2}
%     Given a graph $\Gbf = (\Xbf, (\CV, \CE))$ and $L+1$ graph encoders $\{\hbf^{(l)}_{\Theta}(\cdot)\}_0^L$, the objective is to identify a personalized encoder $l^*(v) \in [0, L]$ for each node $v \in \CV$ such that $\hbf^{(l^*(v))}(v)$ best predicts the true label $y(v)$. 
% \end{problem}
% }
% \ks{propose the problem and show the following as effects of learning personalized scope in the experiments. remove from problem and introduction}
As detailed in Section~\ref{sec:intro}, this problem of selecting representations is important to study since it allows us to
\begin{enumerate}[itemsep=0.04em] %[wide, labelwidth=!, labelindent=0pt,itemsep=0.1em]
    \item \textbf{Increased accuracy} through personalized node-specific smoothing level. 
    % neighbors of the nodes where their attributes can be exploited. 
    \item \textbf{Enables deeper architectures} by disregarding over-smoothened representations. % if they are over-smoothened. higher-order
    \item \textbf{More robustness} against adversarial attacks as it ignores suboptimal layers.
    % \item \textbf{Faster convergence} of the training process since we only classify the nodes in their optimal representation space instead of separating all the nodes in a single layer. 
    % \item \ks{Optimality of the scope} \textbf{More interpretability} of the node-level predictions as they can be explained using just their optimal scope while disregarding the neighbors at other hops.
    % \item Discovered scopes can be used to \textbf{explain} the prediction. 
    % \item \textbf{Preserves the analytic} simplicity of the GNN, unlike arbitrary aggregation.
    \item \textbf{Plug-and-play} improvement irrespective of the underlying GNN architecture.  
\end{enumerate}

\section{Methodology}\label{sec:methodology}

% \paragraph{Decoding.} Since we want to find the optimal scope in any given GNN, the encoder architecture remains fixed. Instead, we change how labels are decoded from representations at different layers to identify the optimal layer $l^*(v)$ for label prediction. Existing GNNs decode the labels from final layer representations by training an additional neural network $\gbf_{\Phi}$ to maximize $\log (\gbf_{\Phi} \circ \hbf_i^{(L)})_{y_i},$ where $y_i$ is the actual label of $v_i$. However, if we want to leverage multiple representations for decoding the labels, this technique would not be possible.

\begin{figure*}[t]
    \centering
    \includegraphics[width=0.8\textwidth]{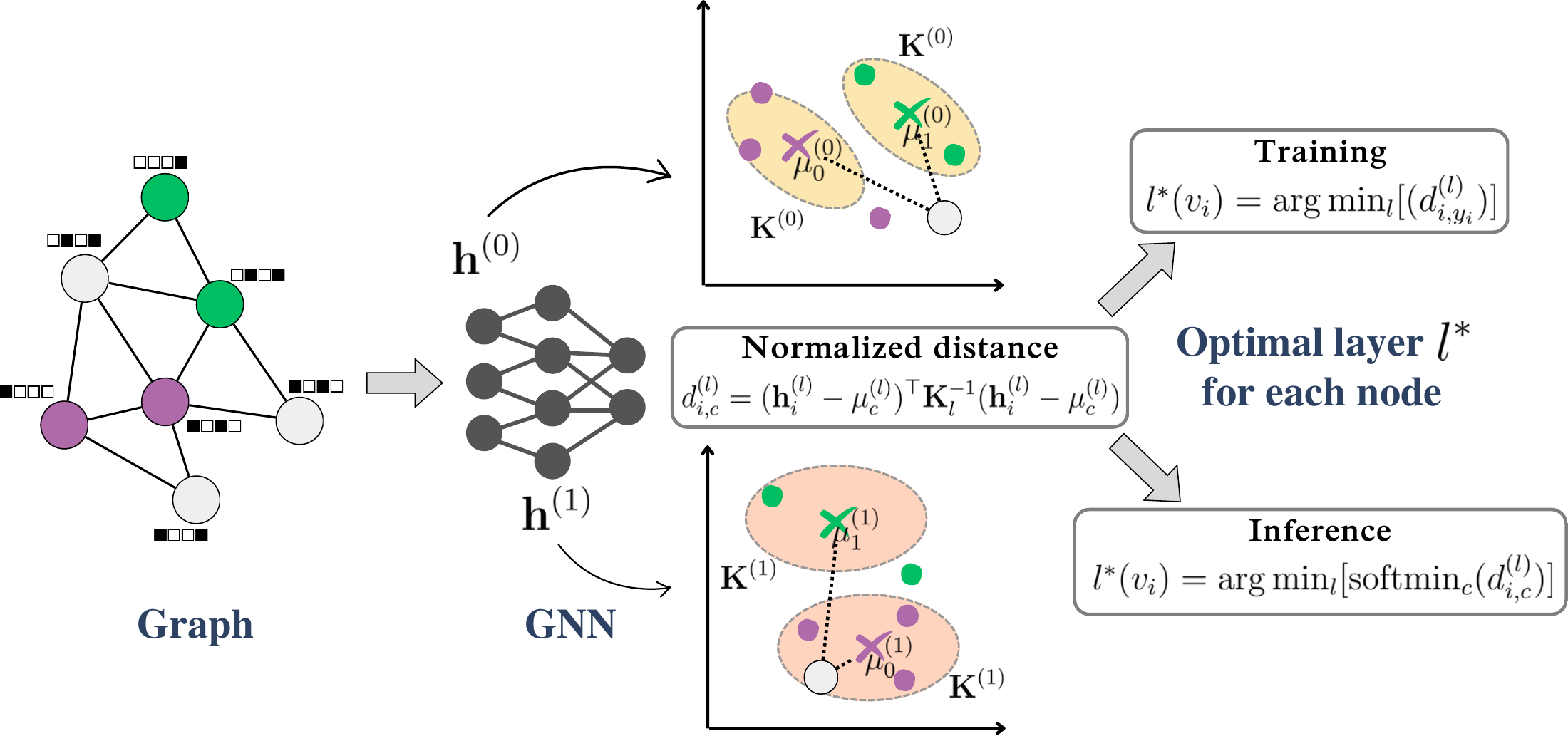}
    \captionof{figure}{GNN with node-optimal prediction layer trained using MetSelect. Labeled nodes are denoted in purple and green while gray denotes unlabeled nodes. 
    % \yd{the inference phase is not introduced in the paper.}
    }
    \label{fig:method}
\end{figure*}

% We assume $\hbf^{(l)}_i \in \mathbb{R}^d$ for all $i \in [1, |\CV|]$ and $l \in [0, L]$ and consider $\hbf^{(0)}_i = \hbf^{0}(\xbf_i)$, where $\hbf^{0}(\cdot): \mathbb{R}^F \rightarrow \mathbb{R}^d$ is either a random transformation or a simple MLP and we chose the latter for more expressivity. 
While the original GNN learns a decoder $\gbf_{\Phibf}(\cdot)$ over $\hbf^{(L)}$ to decode the labels, we compare the $L+1$ representations formed by different layers of the encoder, \ie, $\{\hbf^{(0)}_i, \hbf^{(1)}_i, \cdots, \hbf^{(L)}_i\}$ and pick the one most suitable to classify a given node. We call our method \textbf{MetSelect}\footnote{Code will be open-sourced after publication.}, as we employ \textbf{Met}ric learning to \textbf{select} the node-optimal personalized layer to classify nodes. Figure~\ref{fig:method} illustrates our method with an example of the transductive classification of a graph. 

% form different clusters for each class. we find a prototype representation of each class in each GNN layer and then compare the distance of a node to these prototypes across layers to find the optimal neighborhood scope.

% Then, we choose the layer where the node's representation is closest to any of the label centers. 

% \begin{minipage}[0.5\textwidth]

As noted in Equation~\ref{eq:gnn}, each layer of GNN $\hbf^{(l)}$ represents the nodes in distinct ways (typically, using the $l$-hop neighborhood).
% , \ie, layer $l$. 
Thus, one can separate the nodes based on their labels in each of these spaces. Then, we propose to find the best layer $l^*(v)$ for a node based on how \textit{confidently} that layer predicts its class. To compare the confidence of the prediction from different representation spaces, we adopt \textit{distance}-based class decoding and use the distance as the notion of the confidence~\citep{salakhutdinov2007learning}. However, distance-based decoding can distort the softmax decoding that we do to predict the labels from these layers. Thus, we propose to use a transformed representation $\tilde{\hbf}^{(l)} = \Wbf \hbf^{(l)} + \bbf$ to find the best layer $l^*$. 

% \begin{algorithm}[t]
    
% \end{algorithm}

Distance-based classifiers~\cite{rippel2015metric,deng2019arcface,deng2020sub,salakhutdinov2007learning} typically find a \textit{prototype} embedding of each class and then classify the input to the class whose prototype is the closest in the embedding space. For instance, for layer $l$, the prototype of each class can simply be given as the \textbf{mean} embedding of the training nodes belonging to that class, \ie, 
\begin{equation}\label{eq:mean}
    \resizebox{!}{3.5mm}{%
    $
    \mubf^{(l)}_{c} = \frac{1}{\sum_{v_i \in \CV_{tr}} \one[y_i = c]}\sum_{v_i \in \CV_{tr}} \one[y_i=c] \tilde{\hbf}^{(l)}_i,
    $
    }
\end{equation}
where $y_i$ is the label of the node $v_i$ and $\one[y_i = c] = 1$ if $y_i = c$ and $0$ otherwise. However, we cannot simply compare the distances $\lVert\tilde{\hbf}^{(l)}_i - \mubf^{(l)}_{y_i}\rVert^2$ and $\lVert\tilde{\hbf}^{(l')}_i - \mubf^{(l')}_{y_i}\rVert^2$ where $l' \neq l$ since different spaces can have different class densities. Inspired by \citet{rippel2015metric}, we normalize the distance by the variance to discriminate densities in different spaces. In particular, we find the \textbf{variance} for each space $\tilde{\hbf}^{(l)}$ as simply the empirical variance given the means,
\begin{equation}\label{eq:variance}
    \resizebox{!}{3.5mm}{%
    $
    \Kbf_l = \frac{1}{|\CV_{tr}|-1} \sum_{v_i \in \CV_{tr}} (\tilde{\hbf}^{(l)}_i - \mubf^{(l)}_{y_i}) (\tilde{\hbf}^{(l)}_i - \mubf^{(l)}_{y_i})^\top = \frac{1}{|\CV_{tr}|-1} \sum_{v_i \in \CV_{tr}} (\tilde{\hbf}^{(l)}_i)(\tilde{\hbf}^{(l)}_i)^\top - \sum_{c} \one[y_i = c](\mubf_c^{(l)})(\mubf_c^{(l)})^\top.
    $
    }
\end{equation} 
% since $\sum_{v_i \in \CV_{tr}} \one[y_i = c] \mubf_{c}^{(l)} = \sum_{v_i \in \CV_{tr}} \one[y_i = c] \hbf_i^{(l)}$.
% $\Kbf_l = \frac{1}{|\CV_{tr}|-1} \sum_{v_i \in \CV_{tr}} [\hbf_i^{(l)} {(\hbf_i^{(l)})}^{\top} - \hbf_i^{(l)} {(\mubf_{y_i}^{(l)})}^{\top} - \mubf_{y_i}^{(l)} {(\hbf_i^{(l)})}^{\top} + \mubf_{y_i}^{(l)} {(\mubf_{y_i}^{(l)})}^{\top}]$. 
% $=  \sum_i h_i^{(l)} {(h_i^{(l)})}^{\top} - \sum_c \sum_{i: y_i = c} h_i^{(l)} {(\mubf_{c}^{(l)})}^{\top} - \sum_c \sum_{i: y_i = c} \mubf_{c}^{(l)}{(h_i^{(l)})}^{\top} + \sum_c N_c \mubf_{c}^{(l)} {(\mubf_{c}^{(l)})}^{\top}$, where $N_c = \sum_{i: y_i = c} 1$. But , we can simplify it as $\sum_i [h_i^{(l)} {(h_i^{(l)})}^{\top}] - \sum_c N_c \mubf_{c}^{(l)} {(\mubf_{c}^{(l)})}^{\top}$.
Then, we propose \textbf{MetSelect} that selects the personalized layer $l^*(v)$ such that it minimizes the \textbf{normalized distance} or the Mahlanabois distance, $d_{i,c}^{(l)} = (\tilde{\hbf}^{(l)}_i - \mubf^{(l)}_{c})^\top \Kbf_l^{-1} (\tilde{\hbf}^{(l)}_i - \mubf^{(l)}_{c})$. In particular,
\begin{equation}
\label{eq:metselect_predlayer}
    \resizebox{!}{3.3mm}{%
    $
    l^* (v_i) = \argmin_{l \in [0, L]} \min_{y \in [1, |\CY|]} \left[\exp(-d_{i, y}^{(l)})/\sum_{c=1}^{|\CY|}\exp(-d_{i, c}^{(l)})\right].
    $%
    }
\end{equation}
For training, since we know the true label of each node in the training set, we find the optimal layer as
\begin{equation}
\label{eq:metselectTrain_predlayer}
    \resizebox{!}{3.3mm}{%
    $
    l^* (v_i | y_i) = \argmin_{l \in [0, L]} \left[\exp(-d_{i, y_i}^{(l)})/\sum_{c=1}^{|\CY|}\exp(-d_{i, c}^{(l)})\right].
    $%
    }
\end{equation}
% \textcolor{red}{We note that this can be easily extended to Problem~\ref{prob:main2} to select node-optimal models $\in [0, L]$. }
\paragraph{Training.} Existing GNNs decode the labels from final layer representations by training an additional neural network $\gbf_{\Phi}$ to minimize $\CL_{CE}(v_i, y_i) = \log (\gbf_{\Phi} \circ \hbf_i^{(L)})_{y_i},$ where $y_i$ is the actual label of $v_i$. Since we want to decode from different layers, we employ $L+1$ different decoders $\gbf^{(l)}_{\Phi}$ for $l \in [0, L]$ specialized for each layer. Thus, we train this using a \textit{personalized} Cross-Entropy loss as
\begin{equation}
    \label{eq:loss_metselect_linear}
    % \resizebox{!}{3.5mm}{%
    % $
    \CL_{CE} (v_i, y_i) = - \log {(\gbf^{(l^*(v_i | y_i))}_{\Phi} \circ \hbf_i^{(l^*(v_i | y_i))})}_{y_i}.
    % $
    % }
\end{equation}
% However, this loss does not train the parameters $\Wbf, \bbf$ which form $\tilde{\hbf}$ from $\hbf$. While random transformation matrices were also effective, for more stability, we use an auxiliary loss that minimizes the distance of a training node from the prototype of the correct class in the closest layer $l$ and maximize the distance from the prototypes of other classes in all layers. 
Since learning one decoder per layer can be hard especially for higher number of layers, we also propose to directly leverage the class moments as a distance-based classification. Here, we minimize the distance of a training node from the prototype of the correct class in the closest layer $l$ and maximize the distance from the prototypes of other classes in all layers. Assuming $\alpha=1$, we propose 
\begin{equation}
\label{eq:metselect_distance}
    % \begin{split}
    \resizebox{0.6\hsize}{!}{
    $
    \CL_{\text{distance}} (v_i, y_i) = \max(0, \alpha + d_{i, y_i}^{(l^*(v_i | y_i))} + \log{\sum_{l=0}^L \sum_{j \neq y_i} \exp(-d_{i, j}^{(l)})}),
    % \end{split}
    $
    }
\end{equation}
However, we only use $\CL_{\text{CE}}$ in the experiments unless otherwise mentioned. 

% Both losses allow us to selectively classify the node using only the representation where it is close to the class center and ignore the other layers. 

\begin{minipage}[c]{0.57\textwidth}
    Algorithm~\ref{alg:metselect} describes the training steps in more detail. In particular, we first find the class means and variance for each layer and then use these values to minimize the loss $\CL$ for each training pair $(v_i, y_i): v_i \in \CV_{tr}$. Also, note that we treat the two moments $\mubf_c^{(l)}, \Kbf_l$ as constants in the loss function so that no gradients pass through them for the parameters $\Wbf$ to simplify the gradient descent. \\
    % This is done to simplify the gradient descent step as the moments are updated to reflect the parameter updates in an alternate fashion in every epoch. \\\\
    \textbf{Time Complexity.} The total training time taken per epoch by Algorithm~\ref{alg:metselect} includes the time for (1) Finding the moments of each class cluster, (2) Finding the optimal layer for each node, (3) Calculating the loss over the training nodes, and (4) Updating the parameters using gradient descent. One can find the means $\mubf_c^{(l)}$ in time $O(L|\CY||\CV_{tr}|d)$ and the variances $\Kbf_l$ in time $O(L|\CV_{tr}|d^2)$, where $d$ denotes the embedding dimension. The optimal layers can be found using $O(|\CV_{tr}| L |\CY| d^2)$. One can then decode the probabilities and accumulate the loss in time $O(|\CV_{tr}| |\CY| d)$. 
    This makes up $O(|\CV_{tr}| L |\CY| d^2))$ while the forward pass through the GNN takes at least $O(L d^2)$ (due to multiplication with $d \times d$ weight matrices). Thus, MetSelect remains efficient as long as $|\CV_{tr}|$ and $|\CY|$ are small. 
    % While we empirically find the running time to be similar, specific sampling techniques to reduce the graph size can also be used to scale to larger graphs~\citep{chiang2019cluster}.
\end{minipage}
\hfill
\begin{minipage}[c]{0.4\textwidth}
    \captionof{algorithm}{Training MetSelect}
    \label{alg:metselect}
    \begin{algorithmic}[1]
    % \label{alg:metselect}
        \REQUIRE{{\small GNN $\hbf$ of depth $L$, Training nodes $\CV_{tr} \subset \CV$, Number of epochs $N_e$, Learning rate $\eta$}}
        % , $f_{\Thetabf} = f^{(L)}(f^{(L-1)}(\cdots, f^{(1)}(f^{(0)}(\cdot))))$, $f^{(0)}$ is identity}
        % \STATE Initialize means  and variances $$.
        \STATE $N_c \gets \sum_i \one[y_i = c]$ for all classes $c$.
        \FOR{$e = 1$ to $N_e$}
            \STATE Initialize $\CL \gets 0, \widehat{\mubf}_{c}^{(l)} \gets \zerobf, \widehat{\Kbf}_{l, 0} \gets \zerobf$.
            \FOR{$v_i \in \CV_{tr}$}
                % \STATE Calculate the means {\tiny $\mubf^{(l)}_{c} \gets \sum_{v_i \in \CV_{tr}} \one[y_i=c] \hbf^{(l)}_i/\sum_{v_i \in \CV_{tr}} \one[y_i = c]$} and variances {\tiny $\Kbf^{(l)} \gets \sum_{v_i \in \CV_{tr}} (\hbf^{(l)}_i - \mubf^{(l)}_{y_i}) (\hbf^{(l)}_i - \mubf^{(l)}_{y_i})^\top/(|\CV_{tr}|-1)$}
                % \STATE $\CL \gets \sum_{v_i \in \CV_{tr}} \CL_{\text{MetSelect}}(v_i, y_i; \hbf, \mubf, \Kbf)$ following Equation~\ref{eq:metselect}.
                % \IF{$y_i = c$}
                \FOR{$l \in [0, L]$}
                    \STATE {\small $\widehat{\mubf}^{(l)}_{y_i} \gets \widehat{\mubf}^{(l)}_{y_i} + \hbf^{(l)}_i$.}
                    \STATE {\small $\widehat{\Kbf}_{l, 0} \gets \widehat{\Kbf}_{l, 0} + \hbf^{(l)}_i (\hbf^{(l)}_i)^\top$.}
                \ENDFOR
                % \ENDIF
                \STATE {\small $\CL \gets \CL + \CL(v_i, y_i; l^*)$ [Eqns.~\ref{eq:metselectTrain_predlayer}, \ref{eq:loss_metselect_linear}].} %$,~\ref{eq:metselect_distance}].}
            \ENDFOR
            \STATE $[\Thetabf, \Phibf, \Wbf] \gets [\Thetabf, \Phibf, \Wbf] - \eta \nabla \CL.$  %$\{\zbf^{(l)}\}, y_i)$
            \STATE $\mubf^{(l)}_{c} \gets \widehat{\mubf}^{(l)}_{c}/N_{c}$.
            \STATE {\small $\Kbf_l \gets \tfrac{1}{{|\CV_{tr}| - 1}} (\widehat{\Kbf}_{l, 0} - \sum_{c} {N_c \mubf_c^{(l)} (\mubf_c^{(l)})^\top})$.} 
            % (Eq.~\ref{eq:mean}) and variance $\Kbf^{(l)}$ (Eq.~\ref{eq:variance}) of each class $c$ in layer $l \in [0, L]$. 
        \ENDFOR
    \end{algorithmic}
\end{minipage}

While we empirically find the running time to be similar, specific sampling techniques to reduce the graph size can also be used to scale to larger graphs~\citep{chiang2019cluster}.

\textbf{Space Complexity.} 
% A naive way to calculate the mean and variance would be to store $\hbf^{(l)}$ for each layer and node. However, this can easily blow up the system memory for large graphs. Instead,
As shown in Algorithm~\ref{alg:metselect}, we calculate the means and variance by updating them in an online manner over each node. Thus, the space overhead of MetSelect does not depend on the number of nodes as it only involves $O(L |\CY| d)$ to store the means and $O(L d^2)$ to store the variance. 

\section{Experimental Setup}\label{sec:setup}

\begin{minipage}[c]{0.53\textwidth}
    \textbf{Datasets.} We consider $4$ standard homophilic co-citation network datasets --- Cora, Citeseer, Pubmed~\citep{kipf2016semi} and ogbn-arxiv (ogba)~\citep{hu2020open}, where each node represents a paper that is classified based on its topic area. We also used $6$ heterophilic datasets --- Actor, Chameleon, Squirrel, Cornell, Wisconsin, Texas~\citep{pei2020geom}. Following ~\citet{pei2020geom}, we evaluate the models on $10$ different random train-val-test splits for all the datasets except ogba, where we used the standard OGB split. Table~\ref{tab:datasets} notes the statistics of these datasets along with label homophily, which measures the proportion of nodes with the same labels with an edge in common.
\end{minipage}
\hfill
\begin{minipage}[c]{0.45\textwidth}
% \begin{table}[tb]
    \centering
    \captionof{table}{Statistics of the datasets used in this work. Label Homophily measures the proportion of nodes with same labels that share an edge.}
    \resizebox{1.0\textwidth}{!}{%
    \begin{tabular}{lccccc}
        \toprule
       \multirow{2}{*}{Dataset}  & \# Nodes & \# Edges & \# Features & \# Labels & Label \\
       & $|\CV|$ & $|\CE|$ & $|\Xbf|$ & $|\CY|$ & Homophily\\
       \midrule
       Cora & 2708 & 5429 & 1433 & 7 & 0.83\\
       Citeseer & 3327 & 4732 & 3703 & 6 & 0.71 \\
       Pubmed & 19717 & 44338 & 500 & 3 & 0.79 \\
       Actor & 7600 & 33544 & 931 & 5 & 0.24 \\
       Chameleon & 2277 & 36101 & 2325 & 5 & 0.25 \\
       Squirrel & 5201 & 217073 & 2089 & 5 & 0.22 \\
       Cornell & 183 & 295 & 1703 & 5 & 0.11\\
       Wisconsin & 251 & 499 & 1703 & 5 & 0.16\\
       Texas & 183 & 309 & 1703 & 5 & 0.06 \\
       Ogbn-arxiv & 169343 & 1166243 & 128 & 40 & 	0.65 \\
       \bottomrule
    \end{tabular}}
    \label{tab:datasets}
% \end{table}
\end{minipage}

% \paragraph{Metrics.} Since our objective is a multi-class classification problem, we used different aggregations of the f1-metric, in particular, Accuracy (Acc.) or the micro-f1, class-weighted f1 (f1-wtd), and macro-averaged f1 (f1-mac). 

\textbf{Graph Neural Networks.} We use $3$ representative base GNNs~\footnote{\url{https://pytorch-geometric.readthedocs.io/en/latest/modules/nn.html}} to assess the plug-and-play improvement of our method --- Graph Convolutional Network (GCN)~\citep{kipf2016semi}, Graph Attention Network (GAT)~\citep{velivckovic2017graph}, and Graph Isomorphism Network (GIN)~\citep{xu2018powerful}. All models have depth $L=2$ unless otherwise mentioned.

\textbf{Baselines.} While no baseline exists that directly finds the optimal layer for each node, we extend existing methods for our task to compare them. We use GNN+\{name\} to denote method ``name'' and just GNN to denote GNN trained with FinalSelect.
\begin{itemize}
    \item FinalSelect: Here, we simply use existing GNN implementations that consider the final layer representation for all nodes to predict their labels. 
    \item NDLSelect~\citep{zhang2021node}: NDLS is a node-dependent local smoothing method that trains a simplified GCN~\citep{wu2019simplifying} with varying smoothing levels for each node. We use these node-specific layers $l(v_i)$ to train GNNs using Equation~\ref{eq:loss_metselect_linear} and classify each node accordingly. Note that these personalized layers are agnostic of models and label distribution. 
    \item AttnSelect~\citep{xu2018powerful}: We replace the softmax in JKNet-LSTM to max such that it selects a fixed layer for each node and train using Equation~\ref{eq:loss_metselect_linear} for these node-specific layers. 
\end{itemize}
% \textcolor{red}{We avoid a direct comparison against layer aggregation methods such as JKNet because they do not }
% We often use GNN+MetScope, GNN+NDLScope, and just GNN to denote GNN trained with a personalized scope found by MetScope, NDLScope, and FinalScope for different GNNs.
% Graph Sample and Aggregate (SAGE)~\citep{hamilton2017inductive}, 
% and Simple Graph Convolution (SGC)~\citep{wu2019simplifying}. Since SGC learns a single representation layer with a fixed scope $L$, we consider a set of SGC layers to represent different scopes and select the best one for each node. In particular, for $L=2$, we used two SGCs of depths $1$ and $2$ to train using MetScope. 
% existing MPGnN and the reason we do
% not select other baselines.

\textbf{Implementation.} In GNNs, we assume $\hbf^{(0)} = \Xbf$ but the feature dimension can be very high. Since in MetSelect, we find distances in the embedding space $\hbf^{(l)}$, a high dimension is not suitable due to the curse of dimensionality~\citep{jordan2015machine}. Thus, we used a trainable linear layer to transform the features into a $d$-dimensional space, \ie, $\hbf_i^{(0)} = \Wbf_0 \Xbf_i +\bbf_0 \in \mathbb{R}^d$. We also found other dimensionality reduction methods such as PCA and random transformation to perform similarly. All the models were trained using an Adam optimizer for $500$ epochs with the initial learning rate tuned between $\{0.01, 0.001\}$. The best-trained model was chosen using the validation accuracy and in the case of multiple splits, the mean validation accuracy across splits. All the experiments were conducted on Python 3.8.12 on a Ubuntu 18.04 PC with an Intel(R) Xeon(R) CPU E5-2698 v4 @ 2.20GHz processor, 512 GB RAM, and Tesla V100-SXM2 32 GB GPUs. 
% We also did not do any $\ell_2$ regularization or any other faster convergence methods to train our models as we wanted to compare the effectiveness of the MetScope loss and the Cross-Entropy loss. We found that this did not affect the overall performance of the models while making the comparison fairer. 
% For GraphSAGE, we use the sum aggregation and do not learn root weights for the ogbn-arxiv dataset to obtain the best performance.
% We compare the models trained using MetScope (denoted as GNN+MetScope, \ie, Equation~\ref{eq:metscope}) against the models trained using the standard Cross Entropy loss (denoted as GNN). 

\textbf{Metrics.} Following existing works, we use the \textbf{micro-F1} or \textbf{accuracy} score across different classes for the test set (Test ACC), validation set (Val ACC), and train set (Train ACC) to assess the performance. 
We found that the trends are similar for other aggregations of F1 across different classes such as class-weighted weighted-F1 and macro-F1 scores. 
% \ks{Add one heterophilic GNN}

\section{Results}\label{sec:results}
Here, we show empirically that while solving Problem~\ref{prob:main}, one can make GNNs more \textit{accurate}, \textit{deeper}, and \textit{robust} in a \textit{plug-and-play} manner. 
% Consequently, we assess the effectiveness of our method in enabling these improvements using $6$ different experiments: 
% \textbf{(RQ1)} Can MetSelect be used in a \textit{plug-and-play} manner without reducing the performance?, \textbf{(RQ2)} Does MetSelect \textit{converge} in less number of epochs?, \textbf{(RQ3)} Does MetSelect enable \textit{deeper} GCN models?, \textbf{(RQ4)} Can MetSelect improve performance of GNNs in \textit{heterophilic} datasets?, \textbf{(RQ5)} Are the models trained using MetSelect more robust to poisoning attacks?, and \textbf{(RQ6)} Are the predictions of GNN+MetSelect models self-interpretable?

\subsection{Can MetSelect be used to enhance the performance of GNNs in a \textit{plug-and-play} manner?}

Here, we test the impact of using a personalized prediction layer to classify the nodes from different GNN representations. Table~\ref{tab:all_acc} reports the performance of different layer selection methods on different GNNs and datasets. One can note that our proposed MetSelect method almost always gives the best accuracy among the baselines. This can be observed from the lowest average difference from the maximum accuracy ($\le$ 0.015). Standard GNNs that only use the final layer to classify all nodes are known to obtain good performance in homophilic datasets~\citep{hu2020open}. In these cases, the three personalized layer selection methods (NDLSelect, AttnSelect, and MetSelect) preserve
% {\color{red}[VR: see my comments under Table 2]} 
the performance of the test set across different GNNs for these datasets, with MetSelect being the most consistent among the three. 

GNNs trained using MetSelect can ignore a node's neighbors at a scope that is not useful for its classification. This explains the overall boost in performance on heterophilic datasets. Table~\ref{tab:all_acc} shows that MetSelect can indeed improve/match the performance of the GNNs with a positive increase in the average performance for GCN, GAT, and GIN. In particular, we find a constant improvement in the extremely label-heterophilic datasets such as Cornell, Texas, and Wisconsin. This is particularly because MetSelect learns to ignore the graph structure when predicting labels for many nodes since the graph structure is not aligned with the label distribution for these datasets. Note that this is identified by MetSelect in an automated manner without specifically encoding this preference in the method anywhere. We also note that performance on homophilic datasets does not show a significant increase, which can be explained due to the fact that MetSelect does not enhance the representation power of these models and selecting the most representative final layer tends to be the most powerful for these datasets. Overall, we find that MetSelect provides the best overall performance, being at least $4$ times closer to the best performance across datasets. 

\begin{table*}[t]
    \centering
    \caption{Test F1-score of the GNN models averaged over different splits (except Ogba) in the datasets with standard deviation in the subscript. MetSelect-max inverts the proposed MetSelect method by selecting the layer that maximizes the distance from the class prototype. 
    % Best values are highlighted in bold. 
    $\Delta$ Max denotes the average difference from the mean maximum value across datasets. 
    % Mean Gain measures the average gain in mean performance by using MetSelect over datasets. Label homophily denotes the proportion of nodes that share an edge and have the same labels. $^*$ Heterophilic GNN results are taken directly from ~\citet{yan2022two}. %\ks{Change the SAGE results.}
    % {\color{red}[VR: the +MetSelect is a bit confusing to the reader. In the graphs when you say +MetSelect, it means you are adding MetSelect loss to GCN, GAT, etc, but in the table this consistency of notation is not followed. Here you just say +MetSelect.]}
    }
    \newcolumntype{g}{>{\columncolor{Gray}}c}
    \label{tab:all_acc}
    \resizebox{1.0\textwidth}{!}{%
    \begin{tabular}{c l g g g g g g g g g g | g}
        \toprule
         % & \multicolumn{3}{c}{Select depends on}  
         % \multirow{2}{*}{GNN} & \multirow{2}{*}{Method}\\
        \rowcolor{white}
         & & Cora & Citseer & Pubmed & Actor & Chameleon & Squirrel & Cornell & Wisconsin & Texas & Ogba & $\Delta$ Max $\downarrow$  \\%\multirow{2}{*}{Mean \\ Gain}\\ %\multirow{2}{*}{$\%$ Gain} \\
         % \cmidrule(lr){2-4}
        % \midrule
        % \rowcolor{white}
        % \multicolumn{2}{c}{MLP} & & $0.71 \pm 0.02$ & $0.72 \pm 0.02$ & $0.83 \pm 0.01$ & $0.35 \pm 0.01$ & $0.48 \pm 0.03$ & $0.30 \pm 0.02$ & $0.66 \pm 0.08$ & $0.77 \pm 0.04$ & $0.69 \pm 0.06$ & $0.47$ & \\
        % \rowcolor{white}
        % \multicolumn{2}{c}{GeomGCN~\citep{pei2020geom}$^*$}  & & & & & $0.32 \pm 0.01$ & $0.60 \pm 0.03$ & $0.38 \pm 0.01$ & $0.61 \pm 0.04$ & $0.67 \pm 0.04$ & $0.67 \pm 0.03$ & & \\
        % \rowcolor{white}
        % \multicolumn{2}{c}{GGCN~\citep{pei2020geom}$^*$} & & $0.88 \pm 0.01$ &  $0.77 \pm 0.01$ & $0.89 \pm 0.00$ & $0.37 \pm 0.01$ & $0.71 \pm 0.02$ & $0.55 \pm 0.01$ & $0.86 \pm 0.06$ & $0.87 \pm 0.03$ & $0.85 \pm 0.04$ & \\     
        \midrule
        \rowcolor{white}
        & FinalSelect & $84.99_{1.43}$ & $74.46_{1.72}$ & $87.11_{0.41}$ & $30.40_{1.01}$ & $68.36_{1.85}$ & $52.78_{1.34}$ & $58.65_{8.26}$ & $61.57_{6.68}$ & $67.03_{7.62}$ & $67.75$ & $2.77$\\
        \rowcolor{white}
        & NDLSelect & $85.15_{1.38}$ & $75.98_{1.70}$ & $86.57_{0.41}$ & $28.53_{1.51}$ & $63.75_{1.96}$ & $51.14_{1.52}$ & $43.24_{6.62}$ & $50.98_{7.10}$ & $58.38_{6.14}$ & $63.35$ & $7.37$\\ 
        \rowcolor{white}
        GCN & AttnSelect & $84.87_{1.13}$ & $73.23_{1.32}$ & $86.28_{0.81}$ & $29.68_{1.73}$ & $58.20_{4.96}$ & $31.35_{2.22}$ & $53.78_{11.71}$ & $61.18_{6.84}$ & $61.89_{7.03}$ & $67.29$ & $7.31$\\
        \cmidrule(lr){2-13}
        % \rowcolor{white}
        & MetSelect (ours) & $84.23_{1.53}$ & $74.44_{1.51}$ & $86.60_{0.44}$ & $30.10_{0.79}$ & $66.28_{2.57}$ & $53.66_{1.74}$ & $68.92_{6.40}$ & $76.47_{6.98}$ & $64.05_{7.86}$ & $66.35$ & \textbf{0.97} \\
        \rowcolor{white}
        & MetSelect-max & $76.74_{2.61}$ & $70.48_{2.13}$ & $84.45_{0.60}$ & $29.39_{1.44}$ & $53.63_{1.70}$ & $37.70_{1.53}$ & $50.27_{10.54}$ & $58.63_{4.54}$ & $66.35_{5.30}$ & $55.28$ & 9.79 \\

        \midrule
        
        \rowcolor{white}
        & FinalSelect & $84.61_{1.72}$ & $74.23_{1.65}$ & $86.68_{0.54}$ & $29.13_{0.94}$ & $64.69_{1.63}$ & $46.75_{1.76}$ & $54.05_{6.37}$ & $60.00_{6.28}$ & $62.43_{7.80}$ & $68.58$ & $5.83$ \\ 
        \rowcolor{white}
        & NDLSelect & $83.48_{1.43}$ & $74.61_{1.65}$ & $85.69_{0.58}$ & $28.68_{1.34}$ & $62.39_{2.42}$ & $44.84_{1.66}$ & $42.16_{7.00}$ & $53.92_{6.81}$ & $60.27_{3.61}$ & $65.58$ & $8.79$\\
        \rowcolor{white}
        GAT & AttnSelect & $71.25_{1.94}$ & $73.99_{1.35}$ & $85.75_{0.42}$ & $32.30_{2.17}$ & $49.71_{2.41}$ & $36.69_{2.65}$ & $70.54_{3.24}$ & $80.78_{4.22}$ & $77.03_{2.92}$ & $58.84$ & $5.26$\\
        \cmidrule(lr){2-13}
        % \rowcolor{white}
        & MetSelect (ours) & $82.80_{1.60}$ & $74.53_{1.93}$ & $85.94_{0.43}$ & $28.99_{1.21}$ & $62.27_{2.59}$ & $48.49_{2.40}$ & $70.27_{4.23}$ & $81.96_{4.60}$ & $73.24_{7.15}$ & $66.25$ & \textbf{1.47} \\
        \rowcolor{white}
        & MetSelect-max & $80.41_{2.29}$ & $70.16_{2.16}$ & $84.72_{0.60}$ & $29.76_{1.20}$ & $57.37_{2.94}$ & $44.60_{1.99}$ & $46.22_{7.69}$ & $53.53_{6.83}$ & $61.22_{4.90}$ & $58.26$ & 10.32 \\

        \midrule
        
        \rowcolor{white}
        & FinalSelect & $81.63_{1.56}$ & $69.55_{2.57}$ & $87.19_{0.53}$ & $27.64_{1.22}$ & $68.95_{2.75}$ & $49.88_{2.83}$ & $51.08_{8.59}$ & $61.18_{5.13}$ & $65.14_{8.59}$ & $68.37$ & $5.62$\\
        \rowcolor{white}
        & NDLSelect & $80.72_{1.52}$ & $71.79_{1.63}$ & $86.36_{0.50}$ & $26.55_{1.05}$ & $67.26_{2.41}$ & $53.46_{2.27}$ & $50.27_{8.57}$ & $54.12_{5.48}$ & $60.54_{6.14}$ & $66.18$ & $6.95$ \\
        \rowcolor{white}
        GIN & AttnSelect & $81.29_{1.62}$ & $71.64_{2.17}$ & $86.57_{0.54}$ & $27.62_{1.96}$ & $61.36_{4.00}$ & $41.58_{4.48}$ & $60.00_{10.80}$ & $66.08_{6.06}$ & $70.27_{6.62}$ & $64.51$ & $5.58$\\
        \cmidrule(lr){2-13}
        % \rowcolor{white}
        & MetSelect (ours) & $80.97_{1.52}$ & $72.65_{2.70}$ & $86.73_{0.43}$ & $30.36_{1.21}$ & $63.29_{2.75}$ & $48.56_{2.49}$ & $69.46_{4.04}$ & $79.02_{3.07}$ & $75.68_{4.41}$ & $66.76$ & \textbf{1.33} \\
        \rowcolor{white}
        & MetSelect-max &  $76.29_{1.81}$ & $68.97_{2.43}$ & $85.01_{0.63}$ & $27.59_{1.36}$ & $56.13_{2.36}$ & $41.37_{1.82}$ & $49.86_{9.50}$ & $59.12_{4.73}$ & $62.97_{5.20}$ & $55.83$ & 10.36 \\

        \bottomrule
    \end{tabular}
    }
\end{table*}

\begin{figure*}[t]
    \centering
    \includegraphics[scale=0.7]{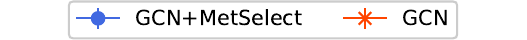} \\
    \hspace*{\fill}
    \subfloat[Cora]{
        \label{fig:nlayers_cora_gcn}
        \includegraphics[width=2.7cm,height=2.5cm]{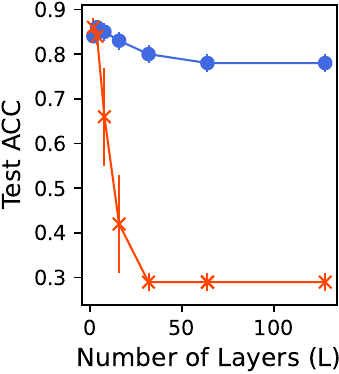}
    }\hfill
    \subfloat[Citeseer]{
        \label{fig:nlayers_citeseer_gcn}
        \includegraphics[width=2.7cm,height=2.5cm]{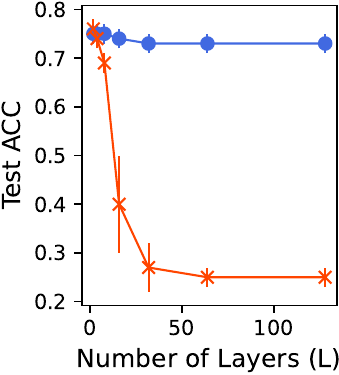}
    }\hfill 
    \subfloat[Pubmed]{
        \label{fig:nlayers_pubmed_gcn}
        \includegraphics[width=2.7cm,height=2.5cm]{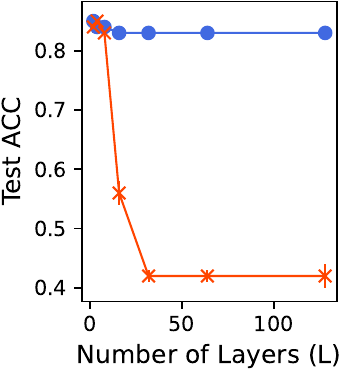}
    } 
    \hspace*{\fill} 
    \caption{Test Accuracy of GCN and GCN+MetSelect with varying depths for different datasets. 
    % \ks{Add ogbn-arxiv}
    }
    \label{fig:nlayers_gcn}
\end{figure*}

\begin{figure*}[t]
    \centering
    \includegraphics[scale=0.7]{Plots/legend.pdf} \\
    \hspace*{\fill}
    \subfloat[Cora Train]{
        \label{fig:mettackTrain_cora_gcn}
        \includegraphics[width=2.7cm,height=2.5cm]{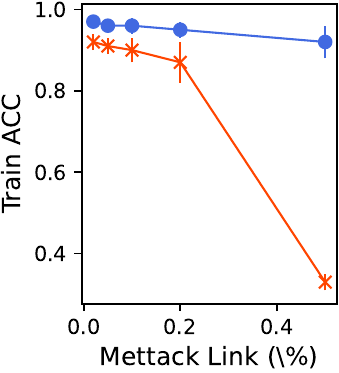}
    }\hfill
    \subfloat[Cora Test]{
        \label{fig:mettackTest_cora_gcn}
        \includegraphics[width=2.7cm,height=2.5cm]{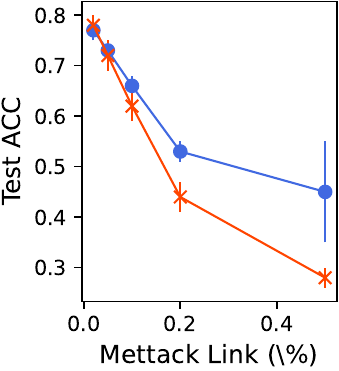}
    }\hfill
    \subfloat[Citeseer Train]{
        \label{fig:mettackTrain_citeseer_gcn}
        \includegraphics[width=2.7cm,height=2.5cm]{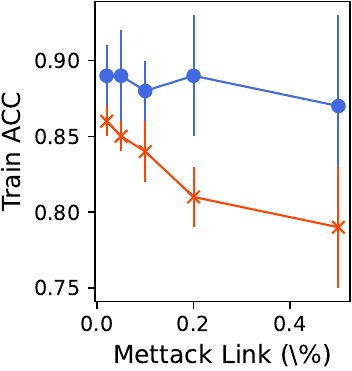}
    } \hfill
    \subfloat[Citeseer Test]{
        \label{fig:mettackTest_citeseer_gcn}
        \includegraphics[width=2.7cm,height=2.5cm]{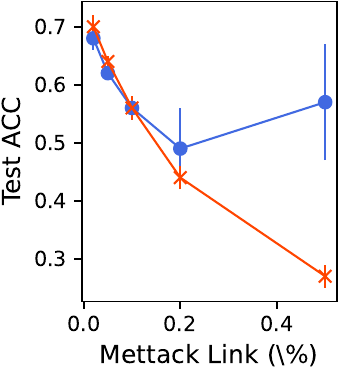}
    } 
    \hspace*{\fill} 
    \caption{Accuracy of GCN and GCN+MetSelect after different strengths of Mettack perturbations.}
    \label{fig:poison_gcn}
\end{figure*}

\subsection{Does MetSelect enable \textit{deeper} GCN models?}
To test MetSelect's effect at different depths, we use the distance-based loss function in Equation~\ref{eq:loss_metselect_linear} to efficiently train the model at larger depths but find similar results to hold for the linear loss (\ie, Equation~\ref{eq:loss_metselect_linear}) till depth $L=8$. 
Since we optimally choose the prediction layer to classify each node, we expect the performance of GCN+MetSelect models to preserve the test performance even when we increase the number of GNN layers. Figure~\ref{fig:nlayers_gcn} demonstrates this effect on Cora, Citeseer, and Pubmed datasets. We find that the accuracy of GCN+MetSelect drops by only up to $10$ accuracy points even for a depth as high as $128$, while the standard GCN model drops by at least $50$ points with just $32$ layers. This shows that MetSelect can indeed enable much deeper GCNs, thus increasing its representative power. 
% \yd{question: will MetSelect select different layers as the optimal layer as we increate the model depth?}\ks{not necessarily} 
% \ks{ogbn-arxiv}
For ogbn-arxiv, we found that while GCN's accuracy reduces from $0.67$ to $0.22$ and $0.06$ at a depth of $16$ and $32$, while MetSelect preserves the accuracy with a slight reduction from $0.66$ to $0.65$ and $0.62$. 
% ; models of depth $> 16$ produced an out-of-memory error on our system due to the term of $O(L |\CV_{tr}|d)$ being very high in the space complexity of MetSelect training (refer Section~\ref{sec:methodology}). 
% We defer the results on other datasets to Appendix~\ref{app:add_exps}. 

\subsection{Are the GNNs trained using MetSelect more \textit{robust} to poisoning attacks?}\label{sec:robust}
% \ks{Here reset?}
Untargeted poisoning attacks~\citep{zugner2018adversarial,sun2020non} have demonstrated that the mean GNN performance can suffer heavily from a change in the graph structure during training. In Section~\ref{sec:problem}, we have highlighted that MetSelect can potentially make the models more robust to such untargeted attacks by ignoring the neighborhood scope not useful for a node's prediction and, thus, potentially the perturbations. To test this, we use Mettack~\citep{zugner2018adversarial} to perturb our training graph that is used to train a GCN model using the standard and the proposed MetSelect loss. In particular, we use the black-box version of Mettack, which finds the attacks using a surrogate GCN model trained on the same dataset. These attacks are structural and represent the flip of a link, \ie, add the link if it doesn't exist and delete it otherwise. The total number of links flipped, or the \textit{budget} is fixed as a parameter that we set to be $p \times |\CE|$ where we vary $p \in [0, 0.5]$ denoted by Mettack Link (\%). 

Figure~\ref{fig:poison_gcn} shows the robustness of the GCN model for Cora and Citeseer at different budgets using just the final layer (denoted as just GCN) and using personalized layer trained using Equation~\ref{eq:metselect_distance} (denoted as GCN+MetSelect). We find that the GCN trained using MetSelect is significantly more robust than the standard GCN with the training accuracy of GCN+MetSelect remaining unaffected by the attacks introduced during training. For instance, while the final layer fails to learn an accurate classifier even for the training set at higher budgets of $50\%$, MetSelect is able to still learn the representations to classify the training set. This is also reflected in the improved generalization of these classifiers as the test accuracy of the GCN+MetSelect is also significantly higher than GCN after the attacks. In particular, 
% \yd{used multiple times, replace it with a synonym}
we see an improvement of up to $30$ points in the test accuracy of GCN trained using MetSelect over the standard GCN. This establishes the robustness of the MetSelect loss owing to its variable prediction layer for nodes.
% and metric learning schema.
% ~\citep{rippel2015metric}. We defer results on other models to Appendix~\ref{app:add_exps}.
% \yd{refer to appendix for additional experiment results}

\begin{figure*}
    \centering
    \hspace*{\fill}
    \subfloat[GCN]{\includegraphics[width=.31\textwidth]{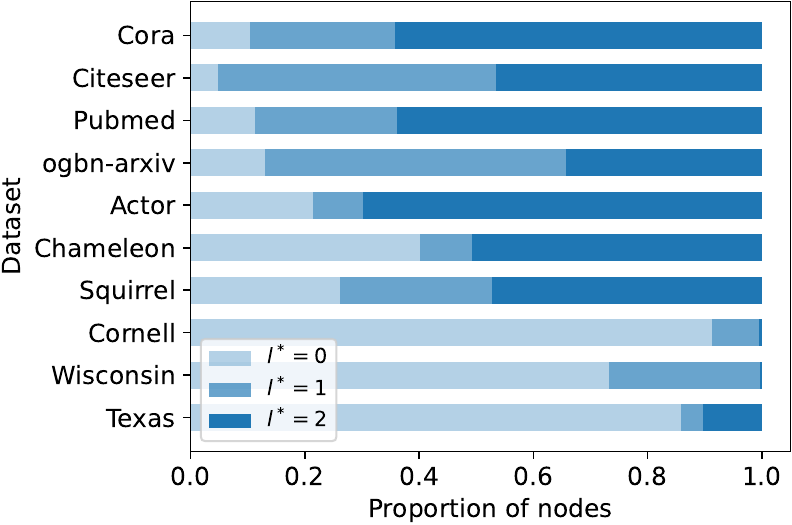} } \hfill
    \subfloat[GAT]{\includegraphics[width=.31\textwidth]{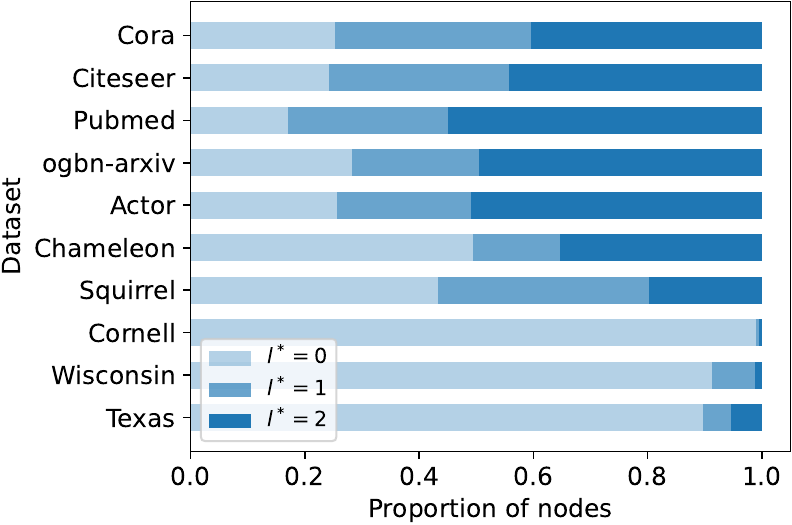} } \hfill
    \subfloat[GIN]{\includegraphics[width=.31\textwidth]{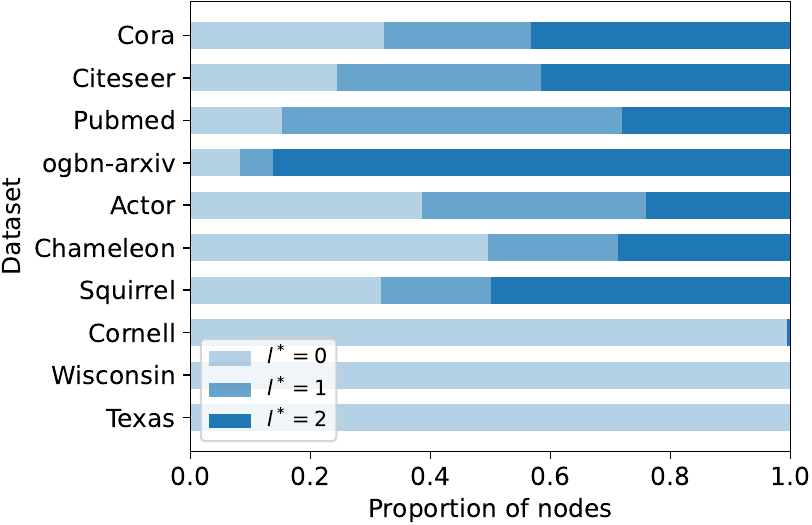} }
    \hspace*{\fill}
    \caption{Proportion of test nodes with a particular personalized layer $\ell^* \in [0, 1, 2]$ as identified by MetSelect.}
    \label{fig:prop_layer_gcn}
\end{figure*}
% \begin{figure*}
%     \centering
%     \hspace*{\fill}
%     \subfloat[GCN]{\includegraphics[width=.47\textwidth]{Plots/Layers/Test.pdf} } \hfill
%     \subfloat[GAT]{\includegraphics[width=.47\textwidth]{Plots/Layers/gat_Test.pdf} }
%     \hspace*{\fill}
%     \caption{Proportion of test nodes with a particular personalized scope as identified by MetSelect.}
%     \label{fig:prop_layer_gcn}
% \end{figure*}

% \begin{table*}[!t]
% \begin{minipage}[c]{0.47\textwidth}
%     \centering
%     \includegraphics[width=\textwidth]{Plots/Interpretability/gcn_Test ACC_h.pdf} 
%     \captionof{figure}{Test Accuracy of GCN+MetSelect with and without the discovered personalized scope $l^*(v)$ for each node $v$. ``Original'' denotes the mean test accuracy using the full scope.}
%     \label{fig:interp_gcn_acc}
% \end{minipage}%
% \hfill
% \begin{minipage}[c]{0.47\textwidth}
%     \centering
%     \includegraphics[width=\textwidth]{Plots/Layers/Test.pdf} 
%     \captionof{figure}{Proportion of test set nodes that have a particular personalized scope $l^*$ for GCN+MetSelect in each dataset.}
%     \label{fig:prop_layer_gcn}
% \end{minipage}
% \end{table*}
% \begin{figure}[t]
%     \centering
     
    % \label{fig:prop_layer_gcn}
% \end{figure}

% \subsection{Are the predictions of GNN+MetSelect \textit{self-interpretable}?}
\subsection{Analysis on MetSelect}

\paragraph{Ablation.} Here, we assess how much the node classification performance depends on the prediction layer identified by MetSelect. To quantify this, we compare it with a method that picks the polar opposite layer of what MetSelect would select. In particular, we consider the one that maximizes the normalized distance from the class center. Table~\ref{tab:all_acc} shows the effectiveness of $l^*(v)$ chosen by the MetSelect model. One can observe that MetSelect almost always outperforms the MetSelect-max (only exception being a little reduction in Actor for GAT) with an average positive gain in accuracy of over $9$ F1 points. 
% The major exception is observed in Cornell for GCN and GAT where the MetSelect-max quickly reaches a local minima of the observed high test accuracy in the first few epochs but does not stay at this value at higher epochs and quickly diverges. Thus, even though the early stopping criterion gives us a high test accuracy, the trained model is not particularly robust. Overall, we observe that MetSelect is the most reliable personalized smoothing method, where we pick the scope that minimizes the distance of the node's representation to the class center instead of maximizing it. 
% {[\color{red}VR: difficult to observe the mse in the figure, either avoid exact numbers, or represent this on the figure.]}\ks{now check.} 
% We also conducted a 2-sample t-test to test the null hypothesis that the accuracy of the ``only $\ell^*$'' model is the same as ``without $\ell^*$'' model on average. We find significant evidence to reject this hypothesis with a p-value of $1.2 \times 10^{-5}$ and the normalized difference between the two models (t-value) being $4.49$. Thus, GCN+MetSelect predictions can be explained by the discovered scope. We provide the results for other models in Appendix~\ref{app:add_exps}.

% As noted in Equation~\ref{eq:metselect_predlayer}, MetSelect finds an optimal GNN representation layer for each node that solves Problem~\ref{prob:main}, in turn, explaining the prediction for a node classification task. Thus, we use the discovered scope $\ell^*$ to assess if the original predictions can be explained using this scope.
% \textcolor{red}{
\paragraph{Self-loops.} We further study the effect of including the feature information in the GNN and see if the gains remain consistent even in this setting. Particularly, we include the node features by adding a self-loop on each node which ensures that the node information is accounted for in all the aggregation layers~\citep{hamilton2017inductive}. Table~\ref{tab:self_loops} shows that the performance remains consistently higher than the final layer selection baseline while highlighting the plug-and-play benefits of our method. 
% }

\begin{minipage}[c]{0.5\textwidth}
    \paragraph{Running Time.} We theoretically found the running time complexity of MetSelect to be higher than existing GNNs. However, we find empirically that it remains within an acceptable factor of $<2.5$ across different datasets to the original method. Table~\ref{tab:time} shows the mean per-epoch time on the largest dataset (ogbn-arxiv) and the remaining small datasets. As compared to NDLSelect, we remain within reasonable bounds of $<1.5$, while sometimes doing even better. Thus, our method is equivalent to the one-time cost of training a proxy model as is done in NDLSelect.
\end{minipage} %
\hfill
\scalebox{0.8}{%
\begin{minipage}[c]{0.57\textwidth}
    \centering
    \captionof{table}{Mean time taken by different algorithms for Ogba and other small datasets. 
    }
    \label{tab:time}
    \resizebox{1.0\textwidth}{!}{%
    \begin{tabular}{c l c c c c c c c c c c}
        \toprule
         % & \multicolumn{3}{c}{Select depends on}  
         % \multirow{2}{*}{GNN} & \multirow{2}{*}{Method}\\
         & & FinalSelect & NDLSelect & MetSelect\\%\multirow{2}{*}{Mean \\ Gain}\\ %\multirow{2}{*}{$\%$ Gain} \\
         % \cmidrule(lr){2-4}
        \midrule
        \multirow{2}{*}{GCN} & Small & 0.15 & 0.37 & 0.43 \\ 
        & Ogba & 0.59 & 1.07 & 1.41 \\
        \midrule
        \multirow{2}{*}{GAT} & Small & 0.17 & 0.41 & 0.48 \\ 
        & Ogba & 0.81 & 1.22 & 1.05 \\
        \midrule
        \multirow{2}{*}{GIN} & Small & 0.07 & 0.26 & 0.20 \\ 
        & Ogba & 0.59 & 0.98 & 1.50 \\
        \bottomrule
    \end{tabular}
}
\end{minipage}
}

We argue that this does not lead to scaling cost as the running time is reasonably small even for the largest OGB dataset with over 100k nodes.

\paragraph{Optimal layer distribution.} 
To further analyze MetSelect, Figure~\ref{fig:prop_layer_gcn} shows the proportion of the nodes that can be effectively classified by a GCN, GIN, GAT using its $l^*$ layer where $l^* \in \{0, 1, 2\}$. 
% We used the distance-based loss of Equation~\ref{eq:metselect_distance} as we find it to have more enhanced layer separation. 
% \yd{confused about the figure.} 
One can note that datasets like Cora, Citeseer, Pubmed, and ogbn-arxiv use higher-order graph structure for prediction, as noted by the high proportion of nodes with $l^* = 2$. On the other hand, datasets like Cornell, Wisconsin, and Texas benefit from a lower-order prediction (\ie, layer $l^* = 0$). This shows that different datasets may have a different layer distribution that is ideal for predicting their node labels and MetSelect can adaptively find this underlying preference to certain representations in an automated manner. 

\begin{table}[tb]
    \centering
    \caption{Comparison of FinalSelect and MetSelect for GCN with self-loops.}
    \label{tab:self_loops}
    \resizebox{0.9\textwidth}{!}{
    \begin{tabular}{l c c c c c c c c c c | c}
        \toprule
         & Cora & Citseer & Pubmed & Actor & Chameleon & Squirrel & Cornell & Wisconsin & Texas & Ogba & $\Delta$ Max $\downarrow$  \\
         \midrule
        FinalSelect & $86.18_{1.15}$ & $74.71_{1.66}$ & $87.02_{0.45}$ & $30.80_{0.92}$ & $68.62_{1.81}$ & $55.99_{1.02}$ & $58.11_{9.56}$ & $61.96_{7.23}$ & $67.84_{5.76}$ & $69.72$ & $2.68$\\
        MetSelect & $83.72_{1.87}$ & $73.05_{1.69}$ & $86.03_{0.56}$ & $30.99_{1.24}$ & $63.75_{2.68}$ & $51.22_{2.47}$ & $66.22_{6.53}$ & $76.67_{5.10}$ & $71.62_{8.18}$ & $66.77$ & \textbf{1.77}\\
         % Gated-GCN & FinalSelect & $83.20_{1.98}$ & $71.90_{1.89}$ & $88.56_{0.57}$ & $34.34_{0.87}$ & $70.66_{1.89}$ & $55.20_{1.47}$ & $63.24_{5.73}$ & $76.08_{4.51}$ & $73.51_{6.95}$ & $34.54_{nan}$ \\
         % & MetSelect & $81.35_{1.30}$ & $71.25_{2.29}$ & $87.51_{0.63}$ & $34.09_{0.80}$ & $69.96_{2.20}$ & $54.28_{2.15}$ & $65.68_{5.85}$ & $74.31_{7.25}$ & $77.03_{6.53}$ & $34.54_{nan}$ \\
         % GTN & FinalSelect & $85.23_{0.82}$ & $74.31_{1.99}$ & $87.13_{0.55}$ & $35.03_{1.46}$ & $65.13_{1.84}$ & $49.37_{1.91}$ & $65.41_{7.07}$ & $76.67_{4.28}$ & $75.41_{4.31}$ \\
         % & MetSelect & $82.88_{1.26}$ & $73.58_{1.88}$ & $86.37_{0.65}$ & $34.34_{0.77}$ & $55.18_{2.11}$ & $39.97_{1.97}$ & $71.08_{3.83}$ & $81.57_{4.91}$ & $73.78_{6.38}$ & $65.06_{nan}$\\
         \bottomrule
    \end{tabular}
    }
\end{table}
% \iffalse
\begin{table}[tb]
    \centering
    \caption{Comparison of MetSelect and layer aggregation methods in GCN using F1/100.}
    \label{tab:jk}
    \resizebox{0.9\textwidth}{!}{
    \begin{tabular}{l c c c c c c c c c c | c}
        \toprule
         & Cora & Citseer & Pubmed & Actor & Chameleon & Squirrel & Cornell & Wisconsin & Texas & Ogba & $\Delta$ Max $\downarrow$  \\
         \midrule
        JKNet-Max & 0.86 & 0.76 & 0.88 & 0.36 & 0.62 & 0.40 & 0.62 & 0.75 & 0.67 & 0.68 & 0.023\\
        JKNet-Cat & 0.86 &  0.76 &  0.88 &  0.35 &  0.61 &  0.42 &  0.66 &  0.78 &  0.71 &    0.69 & 0.011\\
        JKNet-LSTM & 0.71 & 0.72 & 0.88 & 0.35 & 0.50 & 0.29 & 0.70 & 0.79 & 0.75 & 0.68 & 0.046\\
        JKNet-MetSelect &
        0.84 &  0.74 &  0.88 & 0.36 & 0.58 & 0.35 & 0.67 &  0.79 & 0.75 & 0.67 & \textbf{0.009}\\
         % Gated-GCN & FinalSelect & $83.20_{1.98}$ & $71.90_{1.89}$ & $88.56_{0.57}$ & $34.34_{0.87}$ & $70.66_{1.89}$ & $55.20_{1.47}$ & $63.24_{5.73}$ & $76.08_{4.51}$ & $73.51_{6.95}$ & $34.54_{nan}$ \\
         % & MetSelect & $81.35_{1.30}$ & $71.25_{2.29}$ & $87.51_{0.63}$ & $34.09_{0.80}$ & $69.96_{2.20}$ & $54.28_{2.15}$ & $65.68_{5.85}$ & $74.31_{7.25}$ & $77.03_{6.53}$ & $34.54_{nan}$ \\
         % GTN & FinalSelect & $85.23_{0.82}$ & $74.31_{1.99}$ & $87.13_{0.55}$ & $35.03_{1.46}$ & $65.13_{1.84}$ & $49.37_{1.91}$ & $65.41_{7.07}$ & $76.67_{4.28}$ & $75.41_{4.31}$ \\
         % & MetSelect & $82.88_{1.26}$ & $73.58_{1.88}$ & $86.37_{0.65}$ & $34.34_{0.77}$ & $55.18_{2.11}$ & $39.97_{1.97}$ & $71.08_{3.83}$ & $81.57_{4.91}$ & $73.78_{6.38}$ & $65.06_{nan}$\\
         \bottomrule
    \end{tabular}
    }
\end{table}
% \fi
% \textcolor{red}{%
\subsection{How does MetSelect compare with layer aggregation methods?}
Since layer aggregation methods~\citep{xu2018representation} make the predictions from an aggregated/pooled layer using LSTM, max, or concatenation operators, they cannot solve Problem~\ref{prob:main}. In other words, they cannot select one of the existing GNN representations as a personalized layer to predict the node class. Thus, we avoid such a comparison for fairness since unlike layer aggregation methods, we do not enhance existing models' representation power using non-linear activations. \\
For a fair comparison of the two methods, we create a setup where MetSelect also sees the set of pooled representations. Thus, we consider different layer aggregation strategies as multiple representations and select the best aggregation strategy among $\{\hbf^{(\text{LSTM})}, \hbf^{(\text{CAT})}, \hbf^{(\text{MAX})} \}$. We find in Table~\ref{tab:jk} that MetSelect can be used to select personalized aggregation layers for improved performance than using any of the aggregation. This shows complementary gains of our method in addition to simple GNN layer selection.
\section{Conclusion}\label{sec:conclusion}

Our work has shown that graph neural networks are further empowered by exploiting a personalized prediction layer for each node instead of using the same layer for all nodes. This challenges the common notion that the final layer representation of GNNs should fit all nodes. After formalizing the problem of finding the node-optimal GNN layer, we proposed a novel method called MetSelect, to find the optimal layer for a given node in any given GNN.  
% consider varying scopes for each node to 
In particular, we use the layer that minimizes the distance of that node's representation from the class prototype in that layer. 
% We thus find personalized scope since different nodes may have the closest distance with different layers. 
Results show that such a simple strategy can handle oversmoothing, boost performance, and robustness to attacks. 
We hope our work inspires further investigations into node-personalized training of GNNs, including regularization techniques for our method to learn more robust personalized representations. 
% Scalability is another limitation of our work due to a high space complexity, as found for deeper models on the ogbn-arxiv dataset. 
% Future works can also focus on studying the specificity of our method on the base model as hinted by the unexpected results on GraphSAGE. 
While it is extremely hard to prove that a particular layer is optimal for prediction, new benchmarks can be specifically designed for this task either synthetically or by sampling from real-world distributions. 
We also believe that theoretical investigations into how personalized decoding interacts with the training loss function can help gain useful insights into the convergence specificity of node classification of different nodes. 
% We also believe that o
Our work can also be extended to identifying personalized arbitrary motifs for each node for more fine-grained node classification.
Automated layer selection can also inspire sparse architectures of the mixture of experts for graph neural networks~\citep{shazeer2017outrageously}.
% and attention mechanisms. 

\clearpage
% \balance
% \input{impact-statement}

\bibliography{citations}

\begin{thebibliography}{49}
\providecommand{\natexlab}[1]{#1}
\providecommand{\url}[1]{\texttt{#1}}
\expandafter\ifx\csname urlstyle\endcsname\relax
  \providecommand{\doi}[1]{doi: #1}\else
  \providecommand{\doi}{doi: \begingroup \urlstyle{rm}\Url}\fi

\bibitem[Bodnar et~al.(2022)Bodnar, Di~Giovanni, Chamberlain, Li{\`o}, and Bronstein]{bodnar2022neural}
Cristian Bodnar, Francesco Di~Giovanni, Benjamin Chamberlain, Pietro Li{\`o}, and Michael Bronstein.
\newblock Neural sheaf diffusion: A topological perspective on heterophily and oversmoothing in gnns.
\newblock \emph{Advances in Neural Information Processing Systems}, 35:\penalty0 18527--18541, 2022.

\bibitem[Boutyline \& Willer(2017)Boutyline and Willer]{boutyline2017social}
Andrei Boutyline and Robb Willer.
\newblock The social structure of political echo chambers: Variation in ideological homophily in online networks.
\newblock \emph{Political psychology}, 38\penalty0 (3):\penalty0 551--569, 2017.

\bibitem[Brody et~al.(2021)Brody, Alon, and Yahav]{brody2021attentive}
Shaked Brody, Uri Alon, and Eran Yahav.
\newblock How attentive are graph attention networks?
\newblock \emph{arXiv preprint arXiv:2105.14491}, 2021.

\bibitem[Cai et~al.(2021)Cai, Luo, Xu, He, Liu, and Wang]{cai2021graphnorm}
Tianle Cai, Shengjie Luo, Keyulu Xu, Di~He, Tie-yan Liu, and Liwei Wang.
\newblock Graphnorm: A principled approach to accelerating graph neural network training.
\newblock In \emph{International Conference on Machine Learning}, pp.\  1204--1215. PMLR, 2021.

\bibitem[Chiang et~al.(2019)Chiang, Liu, Si, Li, Bengio, and Hsieh]{chiang2019cluster}
Wei-Lin Chiang, Xuanqing Liu, Si~Si, Yang Li, Samy Bengio, and Cho-Jui Hsieh.
\newblock Cluster-gcn: An efficient algorithm for training deep and large graph convolutional networks.
\newblock In \emph{Proceedings of the 25th ACM SIGKDD international conference on knowledge discovery \& data mining}, pp.\  257--266, 2019.

\bibitem[Chien et~al.(2020)Chien, Peng, Li, and Milenkovic]{chien2020adaptive}
Eli Chien, Jianhao Peng, Pan Li, and Olgica Milenkovic.
\newblock Adaptive universal generalized pagerank graph neural network.
\newblock \emph{arXiv preprint arXiv:2006.07988}, 2020.

\bibitem[Cinelli et~al.(2021)Cinelli, De~Francisci~Morales, Galeazzi, Quattrociocchi, and Starnini]{cinelli2021echo}
Matteo Cinelli, Gianmarco De~Francisci~Morales, Alessandro Galeazzi, Walter Quattrociocchi, and Michele Starnini.
\newblock The echo chamber effect on social media.
\newblock \emph{Proceedings of the National Academy of Sciences}, 118\penalty0 (9):\penalty0 e2023301118, 2021.

\bibitem[Dai \& Wang(2021)Dai and Wang]{dai2021towards}
Enyan Dai and Suhang Wang.
\newblock Towards self-explainable graph neural network.
\newblock In \emph{Proceedings of the 30th ACM International Conference on Information \& Knowledge Management}, pp.\  302--311, 2021.

\bibitem[Deng et~al.(2019)Deng, Guo, Xue, and Zafeiriou]{deng2019arcface}
Jiankang Deng, Jia Guo, Niannan Xue, and Stefanos Zafeiriou.
\newblock Arcface: Additive angular margin loss for deep face recognition.
\newblock In \emph{Proceedings of the IEEE/CVF conference on computer vision and pattern recognition}, pp.\  4690--4699, 2019.

\bibitem[Deng et~al.(2020)Deng, Guo, Liu, Gong, and Zafeiriou]{deng2020sub}
Jiankang Deng, Jia Guo, Tongliang Liu, Mingming Gong, and Stefanos Zafeiriou.
\newblock Sub-center arcface: Boosting face recognition by large-scale noisy web faces.
\newblock In \emph{Computer Vision--ECCV 2020: 16th European Conference, Glasgow, UK, August 23--28, 2020, Proceedings, Part XI 16}, pp.\  741--757. Springer, 2020.

\bibitem[Di~Giovanni et~al.(2022)Di~Giovanni, Rowbottom, Chamberlain, Markovich, and Bronstein]{di2022graph}
Francesco Di~Giovanni, James Rowbottom, Benjamin~P Chamberlain, Thomas Markovich, and Michael~M Bronstein.
\newblock Graph neural networks as gradient flows.
\newblock \emph{arXiv preprint arXiv:2206.10991}, 2022.

\bibitem[Dwivedi et~al.(2023)Dwivedi, Joshi, Luu, Laurent, Bengio, and Bresson]{dwivedi2023benchmarking}
Vijay~Prakash Dwivedi, Chaitanya~K Joshi, Anh~Tuan Luu, Thomas Laurent, Yoshua Bengio, and Xavier Bresson.
\newblock Benchmarking graph neural networks.
\newblock \emph{Journal of Machine Learning Research}, 24\penalty0 (43):\penalty0 1--48, 2023.

\bibitem[Eliasof et~al.(2023)Eliasof, Ruthotto, and Treister]{eliasof2023improving}
Moshe Eliasof, Lars Ruthotto, and Eran Treister.
\newblock Improving graph neural networks with learnable propagation operators.
\newblock In \emph{International Conference on Machine Learning}, pp.\  9224--9245. PMLR, 2023.

\bibitem[Feng et~al.(2022)Feng, You, Wang, and Tassiulas]{feng2022kergnns}
Aosong Feng, Chenyu You, Shiqiang Wang, and Leandros Tassiulas.
\newblock Kergnns: Interpretable graph neural networks with graph kernels.
\newblock In \emph{Proceedings of the AAAI Conference on Artificial Intelligence}, volume~36, pp.\  6614--6622, 2022.

\bibitem[Gandhi \& Iyer(2021)Gandhi and Iyer]{gandhi2021p3}
Swapnil Gandhi and Anand~Padmanabha Iyer.
\newblock P3: Distributed deep graph learning at scale.
\newblock In \emph{15th $\{$USENIX$\}$ Symposium on Operating Systems Design and Implementation ($\{$OSDI$\}$ 21)}, pp.\  551--568, 2021.

\bibitem[Gilmer et~al.(2017)Gilmer, Schoenholz, Riley, Vinyals, and Dahl]{gilmer2017neural}
Justin Gilmer, Samuel~S Schoenholz, Patrick~F Riley, Oriol Vinyals, and George~E Dahl.
\newblock Neural message passing for quantum chemistry.
\newblock In \emph{International conference on machine learning}, pp.\  1263--1272. PMLR, 2017.

\bibitem[Gosch et~al.(2023)Gosch, Geisler, Sturm, Charpentier, Z{\"u}gner, and G{\"u}nnemann]{gosch2023adversarial}
Lukas Gosch, Simon Geisler, Daniel Sturm, Bertrand Charpentier, Daniel Z{\"u}gner, and Stephan G{\"u}nnemann.
\newblock Adversarial training for graph neural networks: Pitfalls, solutions, and new directions.
\newblock In \emph{Thirty-seventh Conference on Neural Information Processing Systems}, 2023.

\bibitem[Gutteridge et~al.(2023)Gutteridge, Dong, Bronstein, and Di~Giovanni]{gutteridge2023drew}
Benjamin Gutteridge, Xiaowen Dong, Michael~M Bronstein, and Francesco Di~Giovanni.
\newblock Drew: Dynamically rewired message passing with delay.
\newblock In \emph{International Conference on Machine Learning}, pp.\  12252--12267. PMLR, 2023.

\bibitem[Hamilton et~al.(2017)Hamilton, Ying, and Leskovec]{hamilton2017inductive}
Will Hamilton, Zhitao Ying, and Jure Leskovec.
\newblock Inductive representation learning on large graphs.
\newblock \emph{Advances in neural information processing systems}, 30, 2017.

\bibitem[Hu et~al.(2020)Hu, Fey, Zitnik, Dong, Ren, Liu, Catasta, and Leskovec]{hu2020open}
Weihua Hu, Matthias Fey, Marinka Zitnik, Yuxiao Dong, Hongyu Ren, Bowen Liu, Michele Catasta, and Jure Leskovec.
\newblock Open graph benchmark: Datasets for machine learning on graphs.
\newblock \emph{Advances in neural information processing systems}, 33:\penalty0 22118--22133, 2020.

\bibitem[Jin et~al.(2020)Jin, Ma, Liu, Tang, Wang, and Tang]{jin2020graph}
Wei Jin, Yao Ma, Xiaorui Liu, Xianfeng Tang, Suhang Wang, and Jiliang Tang.
\newblock Graph structure learning for robust graph neural networks.
\newblock In \emph{Proceedings of the 26th ACM SIGKDD international conference on knowledge discovery \& data mining}, pp.\  66--74, 2020.

\bibitem[Jordan \& Mitchell(2015)Jordan and Mitchell]{jordan2015machine}
Michael~I Jordan and Tom~M Mitchell.
\newblock Machine learning: Trends, perspectives, and prospects.
\newblock \emph{Science}, 349\penalty0 (6245):\penalty0 255--260, 2015.

\bibitem[Kipf \& Welling(2016)Kipf and Welling]{kipf2016semi}
Thomas~N Kipf and Max Welling.
\newblock Semi-supervised classification with graph convolutional networks.
\newblock \emph{arXiv preprint arXiv:1609.02907}, 2016.

\bibitem[Li et~al.(2021)Li, M{\"u}ller, Ghanem, and Koltun]{li2021training}
Guohao Li, Matthias M{\"u}ller, Bernard Ghanem, and Vladlen Koltun.
\newblock Training graph neural networks with 1000 layers.
\newblock In \emph{International conference on machine learning}, pp.\  6437--6449. PMLR, 2021.

\bibitem[Li et~al.(2018)Li, Han, and Wu]{li2018deeper}
Qimai Li, Zhichao Han, and Xiao-Ming Wu.
\newblock Deeper insights into graph convolutional networks for semi-supervised learning.
\newblock In \emph{Proceedings of the AAAI conference on artificial intelligence}, volume~32, 2018.

\bibitem[Luan et~al.(2022)Luan, Hua, Lu, Zhu, Zhao, Zhang, Chang, and Precup]{luan2022revisiting}
Sitao Luan, Chenqing Hua, Qincheng Lu, Jiaqi Zhu, Mingde Zhao, Shuyuan Zhang, Xiao-Wen Chang, and Doina Precup.
\newblock Revisiting heterophily for graph neural networks.
\newblock \emph{Advances in neural information processing systems}, 35:\penalty0 1362--1375, 2022.

\bibitem[Nt \& Maehara(2019)Nt and Maehara]{nt2019revisiting}
Hoang Nt and Takanori Maehara.
\newblock Revisiting graph neural networks: All we have is low-pass filters.
\newblock \emph{arXiv preprint arXiv:1905.09550}, 2019.

\bibitem[Pei et~al.(2020)Pei, Wei, Chang, Lei, and Yang]{pei2020geom}
Hongbin Pei, Bingzhe Wei, Kevin Chen-Chuan Chang, Yu~Lei, and Bo~Yang.
\newblock Geom-gcn: Geometric graph convolutional networks.
\newblock \emph{arXiv preprint arXiv:2002.05287}, 2020.

\bibitem[Rippel et~al.(2015)Rippel, Paluri, Dollar, and Bourdev]{rippel2015metric}
Oren Rippel, Manohar Paluri, Piotr Dollar, and Lubomir Bourdev.
\newblock Metric learning with adaptive density discrimination.
\newblock \emph{arXiv preprint arXiv:1511.05939}, 2015.

\bibitem[Rong et~al.(2019)Rong, Huang, Xu, and Huang]{rong2019dropedge}
Yu~Rong, Wenbing Huang, Tingyang Xu, and Junzhou Huang.
\newblock Dropedge: Towards deep graph convolutional networks on node classification.
\newblock \emph{arXiv preprint arXiv:1907.10903}, 2019.

\bibitem[Rossi et~al.(2024)Rossi, Charpentier, Di~Giovanni, Frasca, G{\"u}nnemann, and Bronstein]{rossi2024edge}
Emanuele Rossi, Bertrand Charpentier, Francesco Di~Giovanni, Fabrizio Frasca, Stephan G{\"u}nnemann, and Michael~M Bronstein.
\newblock Edge directionality improves learning on heterophilic graphs.
\newblock In \emph{Learning on graphs conference}, pp.\  25--1. PMLR, 2024.

\bibitem[Rusch et~al.(2022)Rusch, Chamberlain, Mahoney, Bronstein, and Mishra]{rusch2022gradient}
T~Konstantin Rusch, Benjamin~P Chamberlain, Michael~W Mahoney, Michael~M Bronstein, and Siddhartha Mishra.
\newblock Gradient gating for deep multi-rate learning on graphs.
\newblock \emph{arXiv preprint arXiv:2210.00513}, 2022.

\bibitem[Salakhutdinov \& Hinton(2007)Salakhutdinov and Hinton]{salakhutdinov2007learning}
Ruslan Salakhutdinov and Geoff Hinton.
\newblock Learning a nonlinear embedding by preserving class neighbourhood structure.
\newblock In \emph{Artificial intelligence and statistics}, pp.\  412--419. PMLR, 2007.

\bibitem[Shazeer et~al.(2017)Shazeer, Mirhoseini, Maziarz, Davis, Le, Hinton, and Dean]{shazeer2017outrageously}
Noam Shazeer, Azalia Mirhoseini, Krzysztof Maziarz, Andy Davis, Quoc Le, Geoffrey Hinton, and Jeff Dean.
\newblock Outrageously large neural networks: The sparsely-gated mixture-of-experts layer.
\newblock \emph{arXiv preprint arXiv:1701.06538}, 2017.

\bibitem[Shi et~al.(2020)Shi, Huang, Feng, Zhong, Wang, and Sun]{shi2020masked}
Yunsheng Shi, Zhengjie Huang, Shikun Feng, Hui Zhong, Wenjin Wang, and Yu~Sun.
\newblock Masked label prediction: Unified message passing model for semi-supervised classification.
\newblock \emph{arXiv preprint arXiv:2009.03509}, 2020.

\bibitem[Sun et~al.(2020)Sun, Wang, Tang, Hsieh, and Honavar]{sun2020non}
Yiwei Sun, Suhang Wang, Xianfeng Tang, Tsung-Yu Hsieh, and Vasant Honavar.
\newblock Non-target-specific node injection attacks on graph neural networks: A hierarchical reinforcement learning approach.
\newblock In \emph{Proc. WWW}, volume~3, 2020.

\bibitem[Tang et~al.(2022)Tang, Li, Gao, and Li]{tang2022rethinking}
Jianheng Tang, Jiajin Li, Ziqi Gao, and Jia Li.
\newblock Rethinking graph neural networks for anomaly detection.
\newblock In \emph{International Conference on Machine Learning}, pp.\  21076--21089. PMLR, 2022.

\bibitem[Veli{\v{c}}kovi{\'c} et~al.(2017)Veli{\v{c}}kovi{\'c}, Cucurull, Casanova, Romero, Lio, and Bengio]{velivckovic2017graph}
Petar Veli{\v{c}}kovi{\'c}, Guillem Cucurull, Arantxa Casanova, Adriana Romero, Pietro Lio, and Yoshua Bengio.
\newblock Graph attention networks.
\newblock \emph{arXiv preprint arXiv:1710.10903}, 2017.

\bibitem[Wu et~al.(2019)Wu, Souza, Zhang, Fifty, Yu, and Weinberger]{wu2019simplifying}
Felix Wu, Amauri Souza, Tianyi Zhang, Christopher Fifty, Tao Yu, and Kilian Weinberger.
\newblock Simplifying graph convolutional networks.
\newblock In \emph{International conference on machine learning}, pp.\  6861--6871. PMLR, 2019.

\bibitem[Xu et~al.(2019)Xu, Chen, Liu, Chen, Weng, Hong, and Lin]{xu2019topology}
Kaidi Xu, Hongge Chen, Sijia Liu, Pin-Yu Chen, Tsui-Wei Weng, Mingyi Hong, and Xue Lin.
\newblock Topology attack and defense for graph neural networks: An optimization perspective.
\newblock \emph{arXiv preprint arXiv:1906.04214}, 2019.

\bibitem[Xu et~al.(2018{\natexlab{a}})Xu, Hu, Leskovec, and Jegelka]{xu2018powerful}
Keyulu Xu, Weihua Hu, Jure Leskovec, and Stefanie Jegelka.
\newblock How powerful are graph neural networks?
\newblock \emph{arXiv preprint arXiv:1810.00826}, 2018{\natexlab{a}}.

\bibitem[Xu et~al.(2018{\natexlab{b}})Xu, Li, Tian, Sonobe, Kawarabayashi, and Jegelka]{xu2018representation}
Keyulu Xu, Chengtao Li, Yonglong Tian, Tomohiro Sonobe, Ken-ichi Kawarabayashi, and Stefanie Jegelka.
\newblock Representation learning on graphs with jumping knowledge networks.
\newblock In \emph{International conference on machine learning}, pp.\  5453--5462. PMLR, 2018{\natexlab{b}}.

\bibitem[Yan et~al.(2022)Yan, Hashemi, Swersky, Yang, and Koutra]{yan2022two}
Yujun Yan, Milad Hashemi, Kevin Swersky, Yaoqing Yang, and Danai Koutra.
\newblock Two sides of the same coin: Heterophily and oversmoothing in graph convolutional neural networks.
\newblock In \emph{2022 IEEE International Conference on Data Mining (ICDM)}, pp.\  1287--1292. IEEE, 2022.

\bibitem[Ying et~al.(2018)Ying, He, Chen, Eksombatchai, Hamilton, and Leskovec]{ying2018graph}
Rex Ying, Ruining He, Kaifeng Chen, Pong Eksombatchai, William~L Hamilton, and Jure Leskovec.
\newblock Graph convolutional neural networks for web-scale recommender systems.
\newblock In \emph{Proceedings of the 24th ACM SIGKDD international conference on knowledge discovery \& data mining}, pp.\  974--983, 2018.

\bibitem[Zeng et~al.(2021)Zeng, Zhang, Xia, Srivastava, Malevich, Kannan, Prasanna, Jin, and Chen]{zeng2021decoupling}
Hanqing Zeng, Muhan Zhang, Yinglong Xia, Ajitesh Srivastava, Andrey Malevich, Rajgopal Kannan, Viktor Prasanna, Long Jin, and Ren Chen.
\newblock Decoupling the depth and scope of graph neural networks.
\newblock \emph{Advances in Neural Information Processing Systems}, 34:\penalty0 19665--19679, 2021.

\bibitem[Zhang et~al.(2021)Zhang, Yang, Sheng, Li, Ouyang, Tao, Yang, and Cui]{zhang2021node}
Wentao Zhang, Mingyu Yang, Zeang Sheng, Yang Li, Wen Ouyang, Yangyu Tao, Zhi Yang, and Bin Cui.
\newblock Node dependent local smoothing for scalable graph learning.
\newblock \emph{Advances in Neural Information Processing Systems}, 34:\penalty0 20321--20332, 2021.

\bibitem[Zhang et~al.(2022)Zhang, Liu, Wang, Lu, and Lee]{zhang2022protgnn}
Zaixi Zhang, Qi~Liu, Hao Wang, Chengqiang Lu, and Cheekong Lee.
\newblock Protgnn: Towards self-explaining graph neural networks.
\newblock In \emph{Proceedings of the AAAI Conference on Artificial Intelligence}, volume~36, pp.\  9127--9135, 2022.

\bibitem[Zhao \& Akoglu(2019)Zhao and Akoglu]{zhao2019pairnorm}
Lingxiao Zhao and Leman Akoglu.
\newblock Pairnorm: Tackling oversmoothing in gnns.
\newblock \emph{arXiv preprint arXiv:1909.12223}, 2019.

\bibitem[Z{\"u}gner et~al.(2018)Z{\"u}gner, Akbarnejad, and G{\"u}nnemann]{zugner2018adversarial}
Daniel Z{\"u}gner, Amir Akbarnejad, and Stephan G{\"u}nnemann.
\newblock Adversarial attacks on neural networks for graph data.
\newblock In \emph{Proceedings of the 24th ACM SIGKDD international conference on knowledge discovery \& data mining}, pp.\  2847--2856, 2018.

\end{thebibliography}
\bibliographystyle{tmlr}

% \appendix
% \input{log-appendix}

\end{document}